\RequirePackage{fix-cm}

\documentclass[twocolumn]{svjour3}          
\smartqed  
\usepackage{graphicx}
\usepackage{url}
\usepackage{paralist}
\usepackage{soul, color}
\sethlcolor{green}
\usepackage{amssymb}
\usepackage{xspace}
\usepackage{multirow}
\usepackage{float}
\usepackage{amsmath}
 \usepackage{algorithmic}
 \usepackage{algorithm}
\usepackage{lscape}
\usepackage{xtab}

  \usepackage{soul, color}
  \sethlcolor{green}

\newcommand{\dllite}{DL\mbox{-}Lite}
\newcommand{\dllitea}{DL\mbox{-}Lite_{\mathcal{A}}}

\newcommand{\dlr}{\mathcal{DLR}\xspace}
\newcommand{\dlrifd}{\mathcal{DLR}_{\mbox{\emph{{\fontfamily{phv}\selectfont{\footnotesize ifd}}}}}}

\newcommand{\dlcm}{\ensuremath{\mathcal{DC}\xspace}}

\newcommand{\I}{{\ensuremath{\mathcal{I}}\xspace}}

\newcommand{\logicm}{\vbox to8pt{\vss\hbox{$\mathcal{CFDI}^{\forall-}_\textit{\it nc}$}}}

\newcommand{\ormcfd}{\mbox{ORM2}_{\mbox{\emph{{\fontfamily{phv}\selectfont{\scriptsize cfd}}}}}}

\newcommand{\vsep}{\ensuremath{\,|\,}}

 \journalname{An International Journal}
\begin{document}

\title{Evidence-based lean conceptual data modelling languages
\thanks{This work was partially supported by the National Research Foundation of South Africa and the Argentinian Ministry of Science and Technology.
}
}


\author{Pablo R. Fillottrani     \and 
		C. Maria Keet
}


\institute{Pablo R. Fillottrani \at 
Departamento de Ciencias e Ingenier\'ia de la Computaci\'on, Universidad Nacional del Sur, Bah\'ia Blanca, Argentina and Comisi\'on de Investigaciones Cient\'ificas, Provincia de Buenos Aires, Argentina
              \email{prf@cs.uns.edu.ar}           
              \and
              C. Maria Keet \at 
              Department of Computer Science, University of Cape Town, South Africa
              \email{mkeet@cs.uct.ac.za}           
}


\maketitle

\begin{abstract}
Multiple logic-based reconstructions of conceptual data modelling languages such as EER, UML Class Diagrams, and ORM exist. They mainly cover various fragments of the languages and none are formalised such that the logic applies simultaneously for all three modelling language families as unifying mechanism. This hampers interchangeability, interoperability, and tooling support. In addition, due to the lack of a systematic design process of the logic used for the formalisation, hidden choices permeate the formalisations that have rendered them incompatible.
We aim to address these problems, first, by structuring the logic design process in a methodological way. We generalise and extend the DSL design process to apply to logic language design more generally and, in particular, by incorporating an ontological analysis of language features in the process.
Second, we specify minimal logic profiles availing of this extended process, including the ontological commitments embedded in the languages, of evidence gathered of language feature usage, and of computational complexity insights from Description Logics (DL). The profiles characterise the essential logic structure needed to handle the semantics of conceptual models, therewith enabling the development of interoperability tools. 
There is no known DL language that matches exactly the features of those profiles and the common core is small (in the tractable DL $\mathcal{ALNI}$). 
Although hardly any inconsistencies can be derived with the profiles, it is promising for scalable runtime use of conceptual data models. 

\end{abstract}

\section{Introduction}

Conceptual data models were proposed in the 1970s as a vehicle to describe what has to be stored or processed in the prospective information system or database, aiming to separate those `what' aspects from the design of the `how' to achieve that. 
Many conceptual data modelling languages (CDMLs) have been proposed over the past 40 years by several research communities (notably, originating from relational databases, object-oriented software) and for a range of motivations, such as  
spatial entities in geographic information systems, ontology-driven or not, and aiming for 
simplicity and leanness vs expressiveness. Assessing the modelling features of CDMLs over time, it exhibits a general trend toward an increase in modelling features; e.g., UML  
has identifiers since v2.4.1 \cite{UMLspec12}, ORM 2 has more ring constraints than ORM \cite{Halpin01,Halpin08}, and EER also supports entity type subsumption and disjointness cf. ER \cite{Song09,Thalheim09}. Opinion varies about this feature richness and its relation to model quality \cite{Moody05} and fidelity of capturing all the constraints specified by the customer, and asking modellers and domain experts which features they think they use, actually use, and need showed discrepancies between them \cite{tonesd27}.
From the quantitative viewpoint, it has been shown that advanced features are being used somewhere by someone, but they are used relatively very few times, as both the 
quantitative analysis of CDML feature usage in 168 ORM diagrams \cite{Smaragdakis09}, and 101 EER, UML, and ORM diagrams have shown \cite{KF15er}. 
With such insight into feature usage, it is possible to define an {\em appropriate} logic as underlying foundation of a CDML so as to not only clarify semantics but also use it for computational tasks. 
Logic-based reconstructions open up the opportunity for, among others, automated reasoning over a model to improve its quality (e.g., \cite{Berardi05,Queralt12}) 
and runtime usage of the models, such as conceptual and visual query formulation 
\cite{CKNRS10,Calvanese16,Soylu17} and optimisation of query compilation \cite{Toman11}. 

Concerning the `appropriate' logic, many logic-based reconstructions have been proposed over the years (discussed below), which can be grouped into either the Description Logics (DL)-based approach that propose logics from viewpoint of computational complexity, rather than needs and usages by modellers, or the as-expressive-as-needed approach, such as in full first-order predicate logic. Neither of these proposals, however, has taken a methodological approach to language design
and brush over several thorny details of CDMLs, such as which core types of elements to formalise with their own semantics (aggregation,  association ends), whether to include $n$-aries (when $n \geq 2$, not $n = 2$), and various advanced constraints. This has resulted into an embarrassment of the riches of logic-based reconstructions, which is  hampering the actual use of logic-based conceptual data models in information systems  
and therewith risk sliding into disuse.
These problems with the multitude of incompatible ad hoc formalisations raise the questions of 
\begin{compactenum}[i.]
\item How should one design a logic methodologically?  
\item What would be a compatible set of logics for CDMLs that do take into account model feature usage and ontological commitments of CDMLs?
\end{compactenum}

\begin{figure*}
	\centering
	\includegraphics[width=0.65\textwidth]{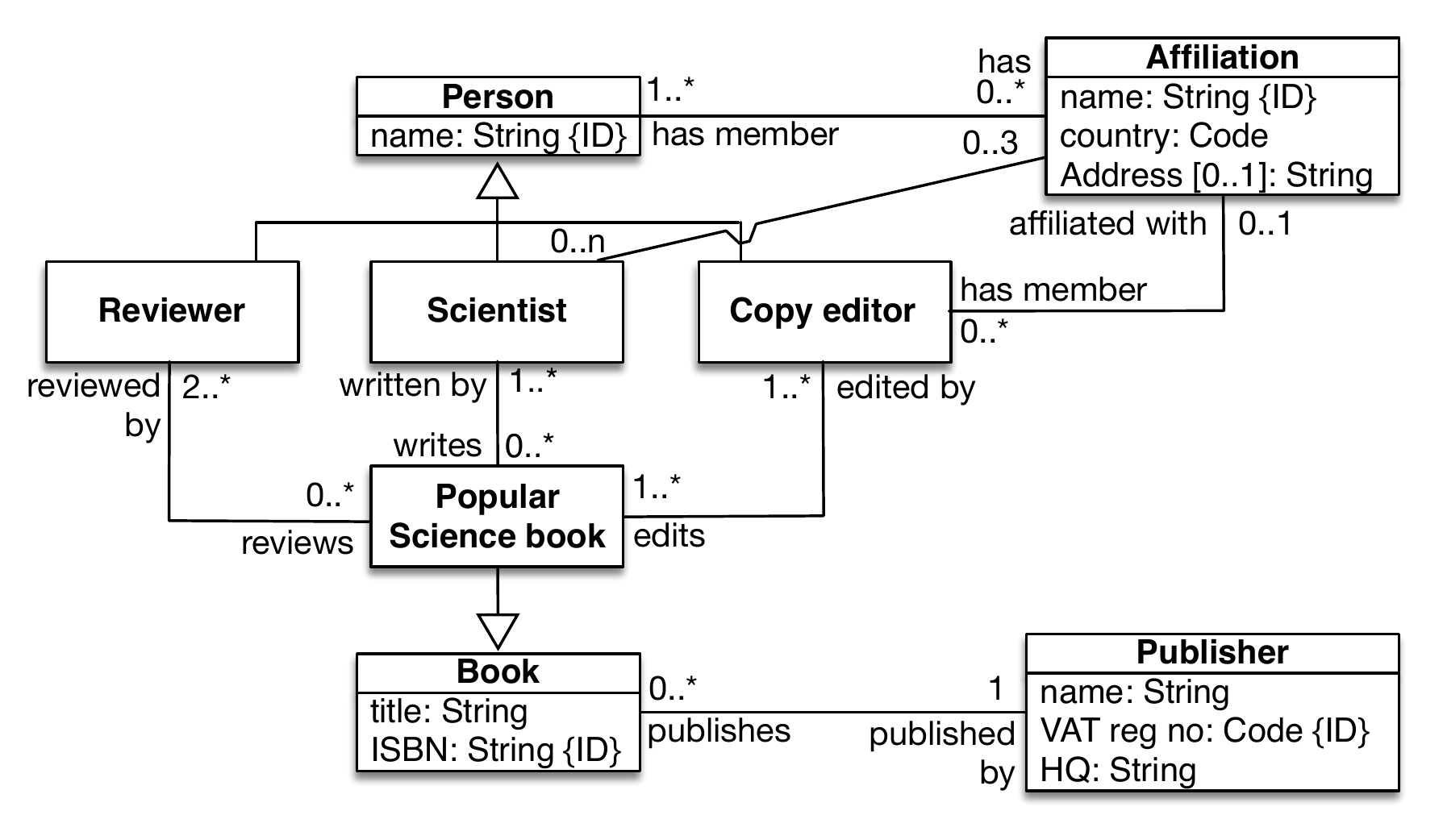}
	\caption{Sample UML Class Diagram containing all possible constraints of the Standard Core Profile, $\dlcm_s$, which emanated from the evidence-based profile specifications.}\label{fig:ExUML}
\end{figure*}

We aim to address these problems and answer these questions in this paper, 
specifically for the structural fragment of the most widely used CDMLs, because this features most prominently as a core interoperability issue in system integration and needs to be resolved before harmonising any `dynamic' components such as methods. 
First, we will adapt and extend Frank's \cite{Frank13} 
domain-specific language (DSL) design methodology into one suitable for language design more broadly, including conceptual data modelling languages, and informed by language design decision points emanating from ontology. Such ontology-driven language design decisions include, among others, positionalism of relations, the conceptual/computational trade-off and 3-dimensionalist vs. 4-dimensionalist. To the best of our knowledge, this is the first inventarisation of parameters of ontological commitments of language design of information and knowledge representation languages. Second, we apply this to the design of logics for several conceptual data modelling languages that is informed by the language feature usage reported in \cite{KF15er}. 
These logic `profiles' cover the most often appearing features used, containing those features that cover about 99\% of the features used in conceptual data models. The outcome is a so-called `positionalist' and a `standard view' core profile, and three language family profiles formalised in a DL, most of which have a remarkable low computational complexity. 
An example of a model in UML Class Diagram notation that can be fully reconstructed into the standard view core profile (more precisely: $\dlcm_s$) is included in Fig.~\ref{fig:ExUML}. It has a logical underpinning thanks to the knowledge base construction rules and three algorithms we propose in this paper, and therewith also has grounded equivalents in EER and ORM notation. 

This paper builds upon several previous papers on this topic \cite{FK15adbis,FK16dl,FKT15,KC16,KF15er}. The main novel contributions in this paper are: i) 
methodological language design, including  ontological aspects involved in it; ii) a new positionalist core profile; 
and 
iii) the profiles have been defined with a formal syntax and semantics. 
The remainder of the paper is structured as follows. The state of the art and related works are discussed in Section~\ref{sec:relworks}. 
Section~\ref{sec:designchoices} presents our first contribution, which is a first inventarisation and discussion of ontological issues that affect language design. Our second main contribution is the logic-based profiles, which are described in Section~\ref{sec:profiles}. 
We close with a discussion in Section~\ref{sec:disc} and conclusions in Section~\ref{sec:concl}. 

\section{Related work}
\label{sec:relworks}

Many conceptual data modelling languages have been proposed over the past 40 years; e.g., UML \cite{UMLspec12}, EER \cite{Chen76,Song09,Thalheim09} and its flavours such as Barker ER 
and IE notation, 
ORM \cite{Halpin89,Halpin08} and its flavours such as CogNIAM and FCO-IM, MADS \cite{Parent06}, and Telos \cite{Mylopoulos90}. Some of those are minor variants in notation, whereas others have to a greater or lesser extent a different number of features.  
Some `families' of languages can be identified, which basically still run along the lines from which subfield it was originally developed. Notably, ER and EER originate from the relational database community \cite{Chen76}, UML Class Diagrams from object-oriented programming \cite{UMLspec12}, and ORM \cite{Halpin08} bears similarities with semantic networks and can be used for conceptual modelling for both database and object-oriented programming, and more recently also business rules. Each `family' has their own set of preferred tools and community of users.

Besides these three main groups, some CDMLs have been developed specifically for additional features (e.g., temporal extensions) or somewhat revised graphical notations of the elements, such as different colours and a so-called `craw's feet' icon vs {\sf ..n} or {\sf ..*} for `many' multiplicity/cardinality. 
We will not address this here, but instead will focus on the underlying language features from a logic-based perspective to which the best graphical elements could be attached as `syntactic sugar' (see, e.g., \cite{Calvanese99,Keet12odcm} for this approach), and language design.

\subsection{Logic-based reconstructions of CDMLs}

A lot of effort has gone into trying to formalise conceptual models for two principal motivations: 1) to be more precise to improve a model's quality and 2) runtime usage of conceptual models. Most works    
are within the scope of 
the first motivation. Notably, various 
DLs have been used for giving the graphical elements a formal semantics and for automated reasoning over them, such as  \cite{Artale07er,Berardi05,Halpin89,Hofstede98,Kaneiwa06,Queralt08}, although also other logics are being used, including OCL \cite{Queralt12}, CLIF \cite{Pan10},  Alloy \cite{Braga10}, and Z \cite{Jahangard11}. There are also different approaches, which are more interested in the verification problem, notably using some variant of constraint programming \cite{Cadoli07,Cabot08}.

Zooming in on DLs, the $\mathcal{ALUNI}$ language has been used for a partial unification of CDMLs \cite{Calvanese99}, whereas other languages are used for particular modelling language formalisations, such as {\em DL-Lite} and $\dlrifd$ for ER \cite{Artale07er} and UML \cite{Berardi05}, and OWL for ORM and UML \cite{Wagih13}. 
These logic-based reconstructions are typically incomplete with respect to the CDML features they cover,  
such as omitting ER's identifiers (`keys') \cite{Calvanese99} or $n$-aries  
\cite{Artale07er,Wagih13}, among many variants. Also, multiple formalisations in multiple logics for one conceptual modelling language have been published. ORM formalisations can be found in, among others, \cite{Franconi12,Halpin89,Hofstede98,Keet09arxiv,Wagih13}, noting that full ORM and ORM2 (henceforth abbreviated as \linebreak `ORM/2') is undecidable due to arbitrary projections over $n$-aries and the acyclic role constraints (and probably antisymmetry). Even for the more widely-used ER and EER (henceforth abbreviated as `(E)ER'), multiple logic- based reconstructions exist from the modeller's viewpoints \cite{Chen76,Song09,Thalheim09} and from the logician's vantage points 
with the $\dlr$ family \cite{Calvanese98,Calvanese99mu,Calvanese01} and $\dllite$ family \cite{Calvanese07} of languages. 

The second principal reason for formal foundations  of CDMLs, runtime usage, comprises a separate track of works, which looks as very lean fragments. 
The driver for language design here is computational complexity and scalability, and the model is relegated to so-called `background knowledge' of the system, rather than the prime starting point for software development. 
Some of the typical runtime usages are: scalable test data generation for verification and validation 
\cite{Nizol14,Smaragdakis09} and ontology-mediated query answering that involves, among others, user-oriented design and execution of queries \cite{Bloesch97,CKNRS10,Calvanese16,Soylu17}, querying databases during the stage of query compilation \cite{Toman11}, 
and recent spatio-temporal \linebreak stream queries that avail of ontology-based data access \cite{KalayciXRKC18,Eiter17,Ozcep15}.

In sum, many logics are used for many fragments of the common CDMLs, where the fragments have been chosen for complexity or availability reasons rather than for which features a modeller actually uses.

\subsection{Language design}
\label{sec:rwld}

\begin{figure*}[t]
\centering
	\includegraphics[width=0.85\textwidth]{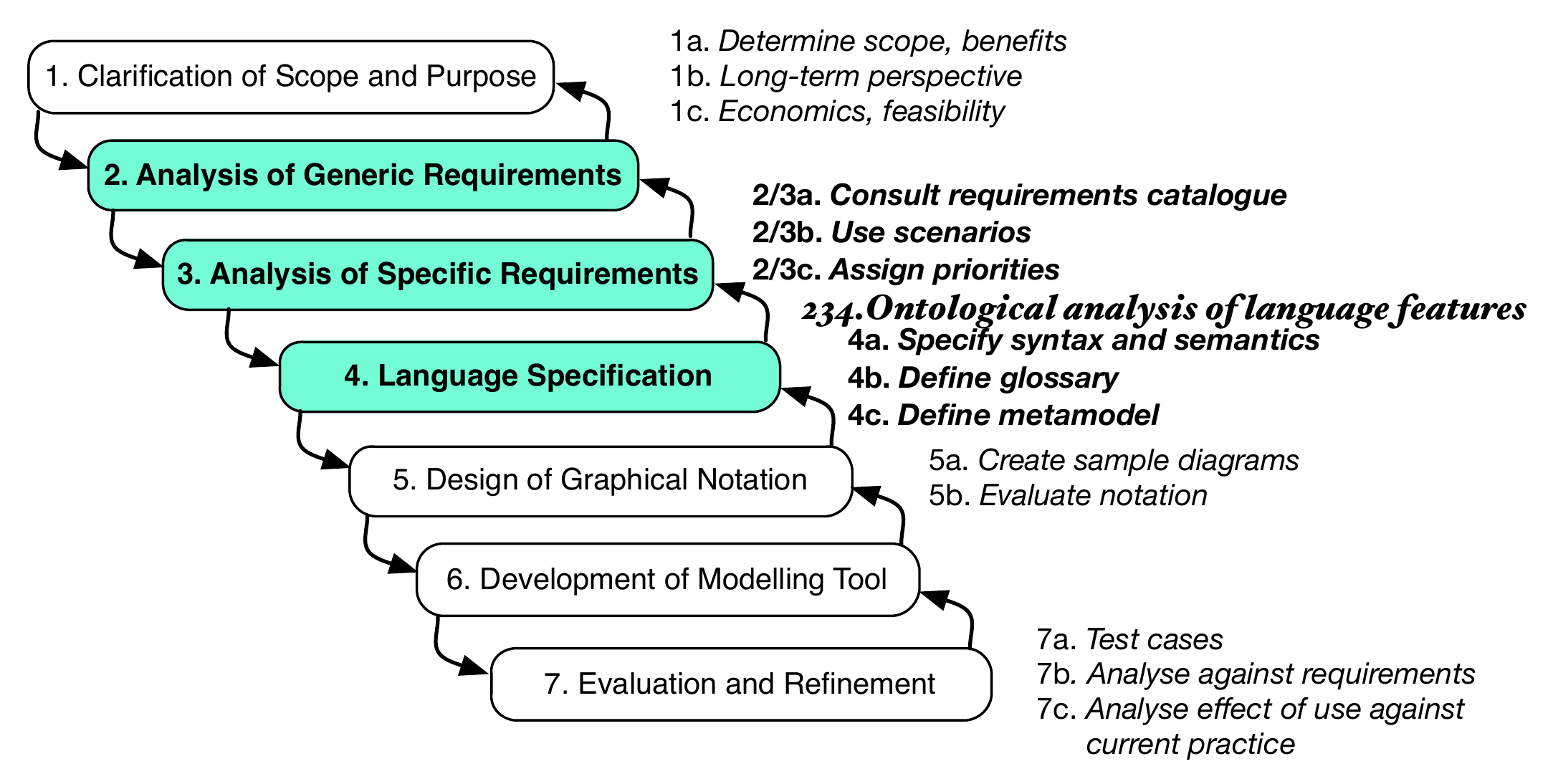}
	\caption{Language design, adapted from \cite{Frank13}, where the focus of this paper is highlighted in bold and shared (green). The ``234. Ontological analysis of language features'' has been added to the 7-step process, which is elaborated on in Section~\ref{sec:designchoices}.}\label{fig:reqeng}
\end{figure*}

The design of a modelling or knowledge representation language may be considered as an engineering task, involving several steps and decision points among alternatives. For instance, a requirement could be that the language has to be decidable in subsumption reasoning or that it needs to have $n$-ary relations, that one can use it for at least the use cases specified upfront, and that it has to have a graphical syntax. Systematic approaches to language design have been developed, especially the design pipeline of Frank \cite{Frank13} and noting others that either focus on one aspect only (e.g., the requirements engineering step \cite{deKinderen15}) or offer guidelines applicable to one class of modelling language specifically (e.g., domain-specific languages \cite{Karsai09}). 
We include here a summary of Frank's waterfall model for domains-specific modelling languages \cite{Frank13}, as shown in Figure~\ref{fig:reqeng}, with the steps we focus on in this paper highlighted. We will step through these stages briefly and with respect to applicability and related works on CDML design.

For step 1,  
the {\em scope} is conceptual data modelling languages, and the {\em goals} are at least model interoperability and, at least to some extent if feasible, runtime use of conceptual models for so-called `intelligent systems' or ontology or conceptual-model driven information systems.

Regarding steps 2 and 3, to the best of our knowledge, there is no requirements catalogue for CDMLs (step 2/3a), but several use cases (step 2/3b) have been reported in scientific literature; e.g., the termbanks for multiple languages requiring interoperability among ER and UML \cite{KF15dke} and runtime integration of information about food in the Roman Empire with ORM \cite{Calvanese16}.
Assigning priorities (step 2/3c) has been done for several languages, but mostly not explicitly. For instance, a priority may be the computational complexity of the problem of satisfiability checking, scalability in the presence of large amounts of data (e.g., the OWL 2 QL modelling language) 
or a scalable model (e.g., OWL 2 EL) \cite{OWL2profiles}, or to have many language features to cover corner cases such as the ``narcissist'' as the single example for the need for local reflexivity (in $\mathcal{SROIQ}$ \cite{Horrocks06}).
Tables with alternative sets of conflicting requirements are available, such as the pros and cons of several logics for formalising conceptual models \cite{FK15adbis} and 
trade-offs for representing the KGEMT mereotopology 
in various logics \cite{KFM12}. For the CDMLs, it has yet to be decided how to assign priorities. 
One could survey industry \cite{Malavolta13}, but it has been shown in at least one survey that modellers do not know the features well enough to be a reliable source \cite{tonesd27}. 
Thus, existing works fall short on providing answers to steps 2 and 3.

There are many papers describing a language specification (step 4), with the DLs the most popular by number of papers. Most of these papers do not have a metamodel, however. Regarding existing metamodels one may be able to reuse for the language specification: there are proposals especially in the conceptual modelling community, spanning theoretical accounts (e.g., \cite{Halpin04,KF15dke}), academic proof-of-concept implementations \cite{Atzeni08,Boyd05,Venable95}, and industry-level applications, such as in the Eclipse Modeling Framework\footnote{\url{https://www.eclipse.org/modeling/emf/}}. The UML diagrams in the OWL and UML standards \cite{OWL2rec,UMLspec12} are essentially metamodels as well. 
To enable a comparison between CDMLs, a unified metamodel is required, which reduces the choice to \cite{Atzeni08,Boyd05,KF15dke,Venable95}. 
The most recent metamodel \cite{KF15dke} 
covers all the static structural components in 
unifying UML Class Diagrams, ER and EER, and ORM and ORM2 at the metamodel layer cf. the subsets in \cite{Atzeni08,Boyd05,Venable95},  
and has both a glossary of elements and the constraints among them. It was  developed in UML Class Diagram notation for communication \cite{KF15dke} and formalised for precision \cite{FK14tr}. 

Step 5 in the language design pipeline---design of a graphical notation---is straightforward for the current scope, for it will be mapped onto the graphical notation of UML Class diagrams, EER, and ORM2. The architecture of the tool (step 6) is 
being worked on \cite{FK16dl,BraunGCF16}, and only after that can one do step 7, which is thus outside of the current scope.

While the 7-step waterfall process for domain-specific languages is generally applicable for logic-based CDML design as well, some ontological analysis during steps 2-4 should improve the outcome, to which we shall return in Section~\ref{sec:designchoices}. 
The case for, and benefits of, using insights from ontology (analytic philosophy) to advance the modelling has been well-documented \cite{Guarino09a,Guarino06}, with many papers detailing improvements on precision of representing the information; e.g., deploying the UFO foundational ontology to improve the UML 2.0 metamodel 
\cite{guizzardi2010} and examining the  
nature of relationships \cite{Keet09orm,Guizzardi08}, and more general philosophical assessments \linebreak about conceptual models, such as regarding concepts \cite{Partridge13} and 3D versus 4D conceptual models \cite{West10}.

Thus, current resources fall short especially on a clear requirements specification and priority-setting for CDMLs and on ontology-driven language design. We will contribute to filling these gaps in the following two sections.

\subsection{Quantitative assessments on language feature use}

To the best of our knowledge, there are only two studies with a quantitative approach to CDML feature usage, which examined ORM diagrams that are used in industry and were developed by one modeller \cite{Smaragdakis09} and examined publicly available EER, UML, and ORM diagrams \cite{KF15er}, 
whose ORM data are similar to those reported in \cite{Smaragdakis09}. 
The diagrams in the dataset of \cite{KF15er} were analysed using a 
unified metamodel \cite{KF15dke}, which facilitated cross-language comparisons and categorisation of entities in those languages into the harmonised terminology.
A relevant selection of the terminology across the languages is included in Table~\ref{table:terminology}.  
This metamodel's top-type is {\sf\small Entity} that has four immediate subclasses: {\sf\small Relationship} with 11 subclasses, {\sf\small Role}, {\sf\small Entity type} with 9 subclasses, and {\sf\small Constraint} that has 49 subclasses (i.e., across the three CDML families, there are 49 different types of constraint). In addition to the hierarchy, the entities have constraints declared among them to constrain their use; e.g., each relationship must have at least two roles and a disjoint object type constraint is only declared on class subsumptions.

\begin{table*}	
\caption{Terminology of the languages considered (relevant selection).}\label{table:terminology}
	\begin{center}
		\begin{tabular}{|| p{2.4cm}| p{3.4cm}| p{2.9cm}|p{3.6cm}|p{2.3cm}||} \hline \hline
{\sf Metamodel term} & \textbf{UML Class Diagram} & \textbf{EER} & \textbf{ORM/FBM} & \textbf{DL} \\ \hline		
			{\sf Relationship} & Association & Relationship & Fact type & Role\\ \hline
			{\sf Role} & Association/ member end & Component of a relationship & Role & Role component \\ \hline
			{\sf Entity type} & Classifier & -- & Object type & -- \\ \hline
			{\sf Object type} & Class & Entity type & Nonlexical object type/ Entity type  & Concept\\ \hline
			{\sf Attribute} & Attribute & Attribute & -- & -- \\ \hline
			{\sf Value type} & -- & -- & Lexical object type/ Value type & --\\ 	\hline
			{\sf Data type} &  LiteralSpecification & -- & Data type & Concrete domain\\ \hline \hline
		\end{tabular}
	\end{center}
\end{table*}

This metamodel was used to classify the entities of the diagrams in a set of 101 UML, (E)ER, and ORM/2 models that were publicly available from online sources, published papers, and textbooks\footnote{the models, their respective provenance, and raw data analysis are available from \url{http://www.meteck.org/swdsont.html}, which is not within the scope of this paper. That experiment design, results, and discussion are described in  \cite{KF15er}.}. The average size of the diagram (vocabulary+subsumption) was found to be about 50 entities/diagram, totalling to 8036 entities, of which 5191 (i.e., 64\%) were entities that were classified in an entity (language feature) that is included in all three language families and 1108 (13.8\%) in ones that are unique to a language family (e.g., UML's aggregation) \cite{KF15er}.
The feature usage results are described and discussed in \cite{KF15er}, where it is noted that while most features of each language family is typically used somewhere, both expected and unexpected high or low frequencies have been observed; e.g., disjoint and covering constraints are used sparingly throughout the models, as are ring constraints in ORM and n-aries or association classes in UML. 
The obtained usage frequency for each entity, together with the design choices described in Section \ref{sec:designchoices}, sustain the logic profiles that will be introduced in Section \ref{sec:profiles}.

\section{Design choices for logic-based profiles}
\label{sec:designchoices}

One could  
take the evidence of which CDML features are used most, and design a logic for it, or pick a logic one likes and make the best out of a logic-based reconstruction. This has resulted in an `embarrassment of the riches' in the plethora of logic-based reconstructions, and even more so when the formalisations are examined in detail, for none is the same. The reason for this is that there are several design choices that each can result in a different logic of different computational complexity, using different reconstruction algorithms, with varying tooling support. Such choices have not been stated upfront, but they have to be piecemeal reconstructed by anyone interested in logic foundations for conceptual models, because such decisions are sparsely discussed in the literature.  
This brings us to the ``4'' of the ``234. Ontological analysis of language features'' extension to Frank's waterfall procedure \cite{Frank13} for language development (recall Fig.~\ref{fig:reqeng} in Section~\ref{sec:rwld}). 
The step 4, ``language specification'', concerns 
affordances and features of the logic, such as 
1) the ability to represent the conceptualisation more or less precisely with more or less constraints\footnote{this is distinct from subject domain coverage; to illustrate the difference: being able to represent, say, ``has part =2 eyes'' vs only ``has part $\geq1$ eyes'' concerns {\em precision}, whereas omitting information about the eyes concerns {\em coverage}.}, and 2) 
whether the representation language contributes to support, or even shape, the conceptualisation and one's data analysis for the conceptual data model for an information system, or embeds certain philosophical assumptions and positions. Regarding the latter, we identified several decision points, which may not yet be an exhaustive list, but it is the first and most comprehensive list to date. They are as follows, and explained and discussed for CDMLs afterward: 
\begin{enumerate} 
	\item Should the CDML be `truly conceptual' irrespective of the design and implementation or also somewhat computational? That is, whether the language should be constrained to permit representation of only the {\em what} of the universe of discourse vs. not only {\em what} but also some {\em how} in the prospective system. The typical example is whether to include data types for attributes or not. 	\label{item:att}
	
	\item Are the roles that objects play fundamental components of relationships, i.e., should roles be elements of the language? 
	\label{item:pos}

	\item Will refinements of the kinds of general elements---that then have their own representation element---result in a different (better) conceptual model? For instance, 
	\begin{enumerate}
		\item to have not just {\sf\small Relationship} but also an extra element for, say,  
		parthood;\label{item:rel} 	
		\item to have not just {\sf\small Object type} but also refinements thereof 
		so as to indicate ontological distinctions between the kind of entities, such as between a rigid object type and the role it plays (e.g., {\sf\small Person} vs {\sf\small Student}, respectively, as being of a different kind);\label{item:otrefine}
		\item if only binary relationships are allowed, the modeller may assume there are only binary relations in the world and reifications of $n$-aries vs the existence of $n$-aries ($n \geq 2$) proper;\label{item:nary}
		\item if only object types are allowed, the modeller may assume everything is a countable object in the world, though one may argue that `stuff', such as {\sf\small Wine} and {\sf\small Gold}, is disjoint from object, and thus would have to be represented with a different element.\label{item:stuff}
	\end{enumerate}
	\label{item:more}  
	\item Does one have a 4-dimensionalist view on the world (space-time worms) and thus a language catering for that or are there only 3-dimensional objects 
	with, perhaps, a temporal extension? 
	\label{item:4d}
	\item What must be named? The act of naming or labelling something amounts to identifying it and its relevance; conversely, if it is not named, perhaps it is redundant.\label{item:5} 
\end{enumerate}
Little is known about what effects the different decisions may have. The data analysis of \cite{KF15er} indicates that binaries vs $n$-aries (Item~\ref{item:nary}) and just plain relationship vs also with aggregation (Item~\ref{item:rel}) does indeed make a difference at least for UML vs (E)ER and ORM/2. UML class diagrams have disproportionately fewer $n$-aries and more aggregation associations than (E)ER and ORM/2. Regarding the former, it is known that $n$-aries in UML class diagrams are hard to read due to the look-across notation \cite{Shoval97} and it uses a different visual element (diamond vs line), compared to (E)ER and ORM/2 that use the same notation for both binaries and $n$-aries; an investigation into the `syntactic sugar' of the graphical elements is outside the current scope. The other options in Item~\ref{item:more} are philosophically interesting, but unambiguous industry requirements for them are sparse, except for `stuff', because quantities of amounts of matter are essential to traceability of ingredients in, among others, the food production process and pills and vaccines production (e.g., \cite{Donnelly10,Solanki16}). In this case, distinguishing between a countable object and a specific quantity of an amount of matter can more precisely relate, say, a specific, numbered, capsule of liquid medicine and the stock that the medicine came from, which, could aid investigations in case of contamination (e.g., the capsule was not cleaned properly vs a contaminated batch of liquid). To date, there is no CDML that includes this distinction other than one proposal for a UML stereotype by \cite{Guizzardi10} and a set of relations for stuffs that could refine UML's aggregation association by \cite{Keet16stuff}.

Continuing with the analysis of Item 3, regarding Object Type refinements (Item~\ref{item:otrefine}), principled ideas have been proposed by \cite{Guarino09a} that avails of notions from philosophy such as rigidity that then leads to types of Object Types, like ``type'' for the kind of objects that supply an identity criterion (the {\sf Person}) and ``role'' that an object plays (the {\sf Student}). This has been applied to UML class diagrams in OntoUML by means of stereotypes \cite{Guizzardi05}.  
More generally, one could take a foundational ontology, such as UFO, GFO, or DOLCE, and use those core categories to refine Object Type (and, analogously, the relationships). 
Choosing one foundational ontology may result in the situation where one's  
conceptual data modelling language---hence, the diagrams---turns out to be incompatible with one that adheres to another foundational ontology \cite{KK13medi}. At the time of writing, it is not clear where or how exactly such refinements provide benefits, other than the typical example of preventing bad hierarchies that OntoClean \cite{Guarino09oc} seeks to resolve. For instance, to not declare, say, {\sf\small Person} as a subclass of {\sf\small Employee}, because the former is a sortal and the latter is a role that the former plays for a limited duration. Instead, it either should be the other way around ({\sf\small Employee} is a {\sf\small Person}) or sideways ({\sf\small Employee} is a role played by {\sf\small Person}). Declaring each Object Type with their respective category and firing ontological rules, such as `a sortal cannot be subsumed by a role', could then detect automatically such quality issues in a diagram. From the viewpoint of logics and formalising it, something like this can be done with second-order logic, many-valued logics, or 
have the elements in the diagrams be subsumed by the entities in a foundational ontology.

The identified decision points are now explained in the following paragraphs.

\paragraph{Conceptual models and features for implementation decisions (Item~\ref{item:att})} 
It is  known theoretically at least that incorporating design and implementation decisions into a model reduces potential for interoperability and integration with other systems. Two common examples are declaring data types of attributes 
(in UML and ORM/2)
and entity type-specific attributes (in (E)ER and UML) vs value types that can be shared among entity types (ORM). ER and EER do not have data types, and its selection is pushed down to the design or implementation stage in the waterfall methodology of database development; e.g., whether some attribute {\sf\small length} should be recorded in {\tt integer} or {\tt float} is irrelevant in the data analysis stage. ORM's Value Types, in contrast to attributes, can be reused by multiple entity types and can easily be reverted into entity types with minimal disruption to the diagram; e.g., a value type {\sf\small length} means the same thing regardless whether it is used for the length of a {\sf\small Sofa}  or a {\sf\small Table} for some furniture database. Of course, one can convert between attributes and value types \cite{FK15adbis}, but a Value Type element in the language enables extra analysis regarding common semantics in a model and constraint specification.

Practically, for the design of a logic, the inclusion of value types entails that one has two types of object types: one that can relate only to other object types and another, disjoint, one that relates to a data type. 
The inclusion of attributes 
affects the language insofar as one wants to create, or already has, a type of relation that relates an object type to a data type, such as 
OWL's Data Property.

\paragraph{Ontological commitments for relationships (Item~\ref{item:pos})}
To determine whether one would want `roles' in the language, let us first illustrate the possible decisions that Item~\ref{item:pos} asks for. Irrespective of the representation decision, there is some relationship, say, {\sf\small teach} that holds between {\sf\small Professor} and {\sf\small Course}. 
One can take the decision of `just predicates', which means that there is assumed to be no {\sf\small teach} relationship, or at least not represented as such, but there would be at least one predicate, {\sf\small teaches} or {\sf\small taught by} in which  {\sf\small Professor} and {\sf\small Course} participate in that specific order or in the reverse, respectively. 
An alternative decision is to choose for the `there are roles too'-option. Here, {\sf\small Professor} plays a role, say, {\sf\small [lecturer]}, in the relationship {\sf\small teach} and {\sf\small Course} plays the role {\sf\small [subject]} in the relationship {\sf\small teach}; hence, role is an element in the language and deemed to be part of the so-called `fundamental furniture' of the universe.  This distinction between predicates-only and roles-too has been investigated by philosophers, and are there called {\em standard view} and {\em positionalist}, respectively. It has been argued that the positionalist commitment with roles (argument places) is ontologically a better way characterising relations than the standard view that enforces an artificial ordering of the elements in the relation and requires inverses\footnote{and anti-positionalist is argued to be better than positionalist \cite{Fine00,Leo08}, but anti-positionalist is unpractical at this stage. A language specific for the anti-positionalist commitment is proposed in \cite{Leo16}.}, and it is also deemed better with respect to natural language interaction and expressing more types of constraints \cite{Fine00,Keet09orm,KC16,Leo08}. 

This has been discussed at some length in \cite{FK15adbis} from a computational perspective, because the main issue is that the CDMLs are all positionalist, yet most logics use predicates with the standard view. 
This is not to say one cannot define roles in, say, FOL, but the notion of role as a component of an $n$-ary predicate is not an element of FOL (it just has predicates, functions, and constants), i.e., FOL adheres to the standard view by design. 
An example of a positionalist logic is $\mathcal{DLR}$ \cite{Calvanese98} and its related logics in the $\mathcal{DLR}$ family, where the syntax does include a ``DL role component'' for the DL roles and it has a corresponding semantics; e.g., an aggregation relationship could then be specified as ${\sf aggregate \sqsubseteq [part]C \times [whole]D}$ where concept {\sf C} plays the role of part in the binary {\sf aggregate} and {\sf D} plays the role of the whole in that relationship\footnote{While some of the $\mathcal{DLR}$ papers assign a number to each role, this was merely for convenience of presentation, not out of necessity, and thus also not to be read as imposing an ordering through the backdoor (Calvanese, pers. comm.).}.

To address the impasse between CDMLs on the one hand and most logics on the other, there are several options. One could commit to a logic-based reconstruction into a positionalist language, such as those in the $\mathcal{DLR}$ family, which are used for partial reconstructions of ER \cite{Calvanese98}, UML class diagrams \cite{Berardi05}, and ORM \cite{Keet09orm}, include roles in the Z formalisation \cite{Jahangard11}, or one that is yet to be designed. One also could deny positionalism  in some way and force it into a standard view logic. For instance, one could change  
the {\sf\small [lecturer]} and {\sf\small [subject]} roles into a {\sf\small teaches} and/or a {\sf\small taught by} predicate and declare them as inverses if both are included in the vocabulary, or pick one of the two and represent the other implicitly through {\sf\small taught by$^-$} or {\sf\small teaches$^-$}, respectively.  
Sampling decisions made in related works showed that, e.g., Pan and Liu 
\cite{Pan10} use a hybrid of both roles and predicates for ORM and its reading labels also may be `promoted' to relationships \cite{Wagih13}, the original ORM formalisation was without roles in the language \cite{Halpin89}, and UML's association ends are sometimes  ignored as well (e.g., \cite{Braga10,Queralt08}), but not always \cite{Berardi05}. 

Exploring the conversion strategies brings one to the computational complexity of the logic. 
Mostly, adding inverses does not change the worst-case computational complexity of a language; e.g., $\mathcal{ALCQ}$ and $\mathcal{ALCQI}$ are both ExpTime-complete under GCIs\cite{Tobies01}.
 A notable exception is the OWL 2 EL profile that does not have inverse object properties \cite{OWL2profiles}. 

\paragraph{3D vs. 4D (Item~\ref{item:4d})} 
The 3-dimensionalist approach takes objects to be 3-dimensional in space where the objects are wholly present at each point in time (i.e., they do not have temporal parts). Statements are true at the `present', whilst being ignorant of the object in the past and future. 
If one wants to deal with time, it is  added as an orthogonal dimension, as in, e.g., \cite{Artale07a,KB17}.
The 4-dimensionalist approach, sometimes also called `fluents', takes entities to exist in four dimensions, being space+time, they are not wholly present at each point in time, and do have temporal parts. Any statements can be about not only the present, but also the past and future (e.g. \cite{Batsakis17,West10}). 
A typical example in favour of 4D is to represent as accurately as possible the notion of a holding or supra-organisation \cite{West10}, such as Alphabet and Nestl\'e: these companies exist for some time and keep their identity all the while they acquire and sell other (subsidiary) companies. In a 3D-only representation, one would have a record of which companies they own now, but not whether they are the same ones as last year (unless temporal information is added in some way; e.g., through database snapshots or time stamps). 

The predominant choice of conceptual data modelling languages is 3D, although temporal extensions do exist especially for ER (e.g., \cite{Artale07a,KB17}). We could not find any evidence that can explain why this is the case.

\paragraph{Naming things (Item~\ref{item:5})}
The act of naming things entails the interaction between natural language and ontology. 
We do not seek to discuss that millennia-old philosophical debate here, but one that is applied within the current context. Naming elements happens differently across the three CDML families. For instance, in UML Class Diagrams, the association ends must be named but the association does not necessarily have to be named, which typically is the other way around in (E)ER except for recursive relations, whereas ORM diagrams commonly only have {\em reading labels} of a relationship rather than naming either the roles or the relationship themselves. ORM thus clearly distinguishes between the conceptual layer and the natural language layer, but such workings are not taken into account explicitly in any of the formalisations, nor has it been explicitly decided whether it should be, other than in a model for roles in \cite{KC16}. 
A systematic solution to this natural language $\leftrightarrow$ ontology interaction has been investigated in the context of the Semantic Web, where the natural language dimension has its own extension on top of an OWL file \cite{Buitelaar14}, although it still relies on naming classes and object properties. What the three CDML families do have in common is that object types are always named. 

For designing a logic, this may not matter much, but it will have an effect on the algorithms to reconstruct the diagrams into a logical theory.

\paragraph{Engineering factors}
As opposed to principles for design with respect to ontological commitments of a logic, there are also practical considerations one may want to take into account. This refers principally to scalability for use of the logic in an intelligent information systems, and the tooling available for the logic for the various support activities, such as an automated reasoner. For instance, (full) first order predicate logic (FOL) is undecidable, unlike the myriad of FOL fragments in the DL family of languages that are typically more computationally well-behaved \cite{Baader08}. 

Automated reasoners offer various services, some of which are more relevant to the context of conceptual data modelling, such as class satisfiability and advanced querying of data, than others (e.g., proving a theorem). Other practical aspects are ease of use of a  tool to design the model (theory), whether a language permits importing other models, and whether other systems are interoperable with it. For instance, OWL 2, by design, allows one to import other ontologies through its {\tt import} statement \cite{OWL2rec}. Even more so with respect tot modular approaches to design, is the Distributed Ontology, Model, and Specification Language DOL\footnote{it was recently standardised by the Object Management Group; \url{http://www.omg.org/spec/DOL/} and \url{http://dol-omg.org/}.}, for it  provides a metalanguage to link models in various logics, including up to second order logic; hence, interoperability with DOL may facilitate its practical use.

	\begin{table*}[t]
		\centering
		\caption{Popular logics for logic-based reconstructions of CDMLs assessed against a set of requirements; ``--'': negative evaluation; ``+'': positive; ``NL-logic'': natural language interaction with the logic; ``OT refinement'': whether the language permits second order or multi-value logics or can only do refinement of object types through subsumption. }
		\begin{tabular}{|p{3.85cm}|p{3.85cm}|p{3.85cm}|p{3.85cm}|}
			\hline 
			\textit{\textbf{DL-Lite}}$_{\mathcal{A}}$ & $\dlrifd$ &  \textbf{OWL 2 DL} &  \textbf{FOL} \\
			\hline \hline
			\multicolumn{4}{|c|}{\em Language features} \\ \hline
			{\footnotesize -- standard view}&{\footnotesize + positionalist} & {\footnotesize -- standard view} & {\footnotesize -- standard view}  \\ \hline		 
			{\footnotesize -- with datatypes}&{\footnotesize -- with datatypes} & {\footnotesize -- with datatypes} & {\footnotesize + no datatypes}  \\ \hline
			{\footnotesize -- no parthood primitive}&{\footnotesize -- no parthood primitive} & {\footnotesize -- no parthood primitive} & {\footnotesize -- no parthood primitive}  \\ \hline
			{\footnotesize -- no $n$-aries}&{\footnotesize + with $n$-aries} & {\footnotesize -- no $n$-aries} & {\footnotesize + with $n$-aries}  \\ \hline		 
			{\footnotesize + 3-dimensionalism}&{\footnotesize + 3-dimensionalism} & {\footnotesize + 3-dimensionalism} & {\footnotesize + 3-dimensionalism}  \\ \hline		
			{\footnotesize -- OT refinement with subsumption}&{\footnotesize -- OT refinement with subsumption} & {\footnotesize -- OT refinement with subsumption} & {\footnotesize -- OT refinement with subsumption}  \\ \hline
			{\footnotesize -- no NL-logic separation}&{\footnotesize -- no NL-logic separation} & {\footnotesize $\pm$ partial NL-logic separation} & {\footnotesize -- no NL-logic separation}  \\ \hline	
			
			{\footnotesize -- very few features; large feature mismatch} & {\footnotesize + little feature mismatch} & {\footnotesize $\pm$ some feature mismatch, with overlapping sets}  
			& {\footnotesize + little feature mismatch}  \\ \hline	
			{\footnotesize -- logic-based reconstructions to complete} & {\footnotesize + logic-based reconstructions exist} & {\footnotesize -- logic-based reconstructions to complete} & {\footnotesize $\pm$ logic-based reconstructions exist} \\ \hline	
			\multicolumn{4}{|c|}{\em Computation and implementability} \\ \hline		
			{\footnotesize + PTIME (TBox); AC$^0$ (ABox)} & {\footnotesize $\pm$ ExpTime-complete} & {\footnotesize $\pm$ N2ExpTime-complete} & {\footnotesize -- undecidable} \\ \hline		
			{\footnotesize + very scalable (TBox and ABox)}&{\footnotesize $\pm$ somewhat scalable (TBox)} & {\footnotesize $\pm$ somewhat scalable (TBox)} & {\footnotesize -- not scalable}  \\ \hline		
			{\footnotesize + relevant reasoners} &{\footnotesize -- no implementation} & {\footnotesize + relevant reasoners} & {\footnotesize $\pm$ few relevant reasoners}  \\ \hline				 		 			 		 
			{\footnotesize + linking with ontologies doable} & {\footnotesize -- no interoperability} & {\footnotesize + linking with ontologies doable} & {\footnotesize -- no interoperability with widely used infrastructures}  \\ \hline			 
			{\footnotesize + `integration' with DOL} & {\footnotesize -- no integration with DOL} & {\footnotesize + `integration' with DOL} & {\footnotesize + `integration' with DOL}  \\ \hline			 			 		  
			{\footnotesize + modularity infrastructure} & {\footnotesize -- modularity infrastructure} & {\footnotesize + modularity infrastructure} & {\footnotesize -- modularity infrastructure}  \\ \hline		 		 		 
		\end{tabular}
		\label{tab:consider}
	\end{table*}

\paragraph{Assessment of popular languages and their commitments}
We assessed the relatively frequently occurring logics for formalising conceptual data models on whether it would be possible to choose a `best fit' language. This comparison is included in Table~\ref{tab:consider}. The first section of the table summarises the main design decisions discussed in the preceding paragraphs, whereas the second part takes into consideration non-ontological aspects with an eye on practicalities. Such practicalities include among others, scalability, tooling support, and whether it would be easily interoperable with other systems. 

The first section in the table suggests $\dlrifd$ and FOL are good candidates for logic-based reconstructions of conceptual data models; the second section has more positive evaluations for $\dllitea$ and OWL 2 DL. Put differently: neither of them is ideal, so one may as well design one's own language for the best fit.

\section{Logic-based profiles for conceptual data modelling languages}
\label{sec:profiles}

We now proceed to 
define logics to characterise 
the model-theoretic semantics such that it is minimalist with respect to the most-used features for each of the three families of CDMLs. 
This takes into account the ontological considerations discussed in the previous section as well as the evidence from \cite{KF15er} and the requirement to have a coverage of around 99\% of the used entities and constraint. Because of afore-mentioned ontological reasons in favour of roles as well as that all three CDML families are positionalist, a Positionalist Core is defined despite its current lack of implementation support (Section~\ref{sec:poscore}). Afterward, a standard view Standard Core and language-specific profiles are defined in Sections~\ref{sec:score}-\ref{sec:ormprofile}.

\begin{figure*}[t]
	\centering
	\includegraphics[width=0.8\textwidth]{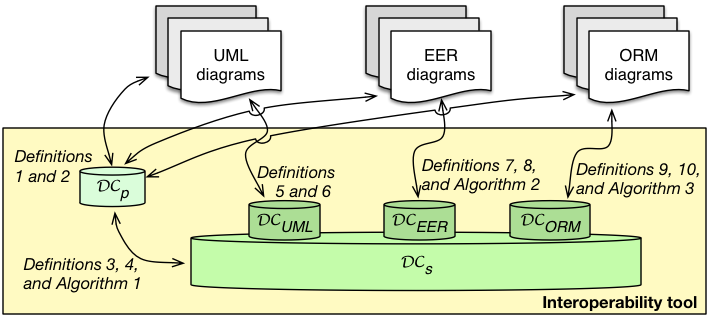}
	\caption{Sketch of the orchestration between the profiles and algorithms.}\label{fig:orchestration}
\end{figure*}

An overview of the upcoming definitions and algorithms is shown in Fig.~\ref{fig:orchestration}.
These profiles constitute a theoretical backbone for an interoperability tool between conceptual model expressed in different graphical languages and with different philosophical assumptions. 
The main distinction is the positionalist or the standard view, resulting in profiles $\dlcm_p$ and $\dlcm_s$, which  each formalise the most widely used language features. The standard view profile is then extended into three different profiles one for each CDML, and serves as background knowledge to be exchanged between profiles. In order to interoperate from the positionalist and the standard view profiles some compromises must be taken, described mainly in Algorithm \ref{alg:eer}. Importing conceptual models into CDML may be carried out also with this theoretical structure, while exporting may be done by translating reasoner output into a suitable textual representation.

\subsection{Profiles}

Positionalism is the underlying commitment of the relational model and a
database's physical schema, as well as of the main CMDLs. It has been employed in Object-Role Modelling (ORM) and its precursor NIAM for the past 40 years \cite{Halpin08}, UML Class Diagram notation requires
association ends as roles, and Entity-Relationship (ER) Models have relationship components \cite{FK15adbis}. On the other hand, First Order Logic and most of its fragments, notably standard DLs \cite{Baader08},  do not exhibit roles (DL role components) 
among its vocabulary. In order to be able  to do reasoning, conceptual schemas written in these CMDLs are generally translated into a DL by removing roles, and thus losing 
the connection that exists when the same role is played by different concepts,
as the following example shows.
\begin{example}
Consider de ER diagram shown in Figure \ref{fig:ERD}, representing a a (partial) conceptual model. In the rent relationship any person may rent any real state property, whereas in the house mortgage relationship any person living in a residential property may put a lien on it to obtain a loan from the bank. Both relationships involve the {\sf occupant} role played by instances of the {\sf Person} entity. 
This role name is relevant for querying, say, the real estate occupants in the database, so it is relevant for the model's intended meaning. 
Following the translation procedure described in \cite{Artale09thedl-lite} to the \dllite{}	
DL family, the role {\sf occupant} in the {\sf rent} relationship is translated as 
$$
\begin{aligned}
\exists \mathsf{occupant} & \sqsubseteq \mathsf{Person}\\
   \exists \mathsf{occupant}^{-} & \sqsubseteq \mathsf{RealStateProperty}  
  \end{aligned}  
$$
The two formula state the domain and the range of the role.
Similarly, the role {\sf occupant} in the {\sf houseMortgage} relationship is translated, including the functional constraint, as
$$
\begin{aligned}
\exists \mathsf{occupant} & \sqsubseteq \mathsf{Person}\\
\exists \mathsf{occupant}^{-} &\sqsubseteq \mathsf{Residential}\\
  \geq 2\, \mathsf{occupant} &\sqsubseteq \bot
 \end{aligned}
$$
In the case that both translations are merged into the same conceptual model formalisation, then unintended meaning is obtained. For example,  that only houses may be rented, and that a person may rent only one property. The solution is to rename one of the roles, but then the connection between both roles is lost in the formalisation, the role is split, and therewith the intended meaning is weakened. This problem does not arise in the translation to  the positionalist $\dlrifd$ following \cite{Berardi05}.
\begin{figure}
	\includegraphics[width=8.2cm]{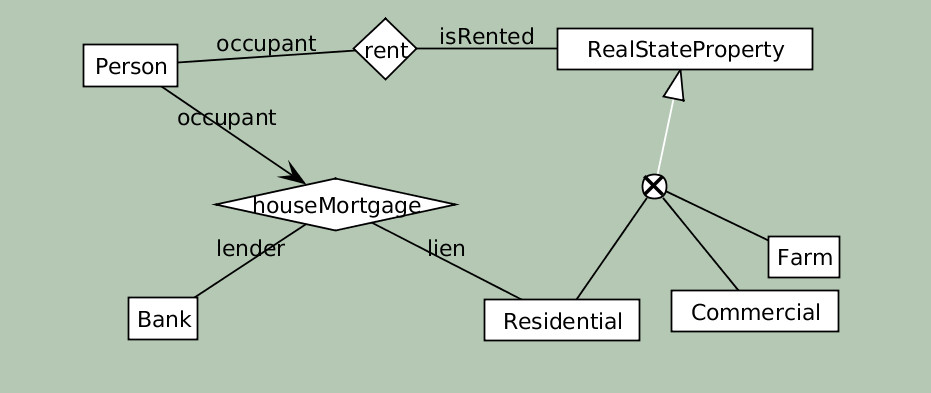}
\caption{ER diagram showing multiple uses of the same role.}\label{fig:ERD}
\end{figure}
\end{example}
Therefore, we consider it relevant to design a positionalist core profile that preserves roles as first class citizens among the DL vocabulary.
In case reasoning over  advanced modelling features is needed, it is possible to switch to the standard core profile with the cost of losing this connection. This translation is given 
in Algorithm~\ref{alg:eer} further below.

\subsubsection{Positionalist Core Profile}
\label{sec:poscore}

In this section we define the DL fragments that describes the positionalist core profile. We use the standard DL syntax and semantics, as given in \cite{Baader08,Berardi05}.
\begin{definition}\label{def:dlcmp}
	Given a conceptual model in any of the analysed CDMLs, we construct a {\em knowledge base} in $\dlcm_p$ by applying the rules:
	\begin{enumerate}
		\item we take the set all of object types $A$, binary relationships $P$, datatypes $T$ and attributes $a$ in the model as the basic elements in the knowledge base.
		\item for each binary relationship $P$ formed by object types $A$ and $B$, we add to the knowledge base the assertions $\geq 1 [1] P \sqsubseteq A$ and 
		$\geq 1 [2] P \sqsubseteq B$.
		\item for each attribute $a$ of datatype $T$ within an object type $A$, including the transformation of ORM's Value Type following the rule given in \cite{FK14}, we 
		add the assertion $A\sqsubseteq \exists a.T \sqcap \leq 1 a$. 
		\item subsumption between two object types $A$ and $B$ is represented by the assertion $A\sqsubseteq B$.
		\item for each object type cardinality $m..n$ in relationship $P$ with respect to its $i$-th  component $A$,
		we add the assertions  $A \sqsubseteq\, \leq n [i] P\, \sqcap \geq m [i] P$. 
		\item we add for each  mandatory constraints of a concept $A$ in a relationship $P$   the axiom $A\sqsubseteq\, \geq 1 [1] P$ or $A\sqsubseteq\, \geq 1 [2] P$
		depending on the position played by $A$ in $P$. This is a special case of the previous one, with $n = 1$.
		\item for each single identification in object type $A$  with respect to an attribute $a$ of datatype $T$ we add the axiom $\mathsf{id}\,A\,a$.
	\end{enumerate}
\end{definition}
This construction is linear in the number of elements in the original conceptual model, so the overall complexity of the process (translation and then reasoning) on the theory is the same as on the conceptual model.
We restrict it to binary relationships only, because general  $n$-ary relationships are rarely used in the whole set of analysed models. The
(E)ER and ORM/2 models exhibit a somewhat higher incidence of $n$-aries,
so they are included in the respective profiles (see below). Also, we allow only one such constraint for each component, as multiple cardinality constraints over the same component in a relationship 
are used very rarely.

$\dlcm_p$ can be represented by the following DL syntax. Starting from atomic elements, we can construct binary relations $R$, arbitrary
concepts $C$  and axioms $X$ according to the rules: 
\begin{align*}
C &\longrightarrow \top \vsep A \vsep \leq k[i] R \vsep \geq k[i] R \vsep \forall a.T \vsep \exists a.T\vsep \leq 1\, a \vsep \\ 
    & \mbox{\hspace{8mm}} C \sqcap D \\
 R &\longrightarrow \top_2 \vsep P \vsep (i:C)  \\
X &\longrightarrow C\sqsubseteq D \vsep \mathsf{id}\, C\, a
\end{align*}
where $i=1,2$ and  $0<k$. For convenience of presentation, we generally use the numbers 1 and 2 to name the role places, but they can be any number or string
and do not impose an order.
Whenever necessary we note with $U$ the set of all role names  in the vocabulary, with $\mathtt{from}, \mathtt{to} \in U$ fixed argument places for attributes such that  
$[\mathtt{from}]$ is the 
role played by the concept, and $[\mathtt{to}]$ the role played by the datatype. These names must be locally unique in each relationship/attribute.

Although this syntax represents all $\dlcm_p$ knowledge bases, there are sets of formula following the syntactic rules that are not $\dlcm_p$ knowledge bases since they are not result of any translation of a valid conceptual model. For example, the knowledge base $\{ A\sqsubseteq \exists a.T \sqcap \leq 1 a\sqcap \forall a.T\}$ is not a $\dlcm_p$ knowledge base, it can't be obtained from the translation of any diagram.

Now we introduce the semantic characterisation. 
\begin{definition}
	An {\em  $\dlcm_p$ interpretation} $\cal{I}=(\Delta_C^{\cal{I}},\Delta_T^{\cal{I}},\cdot^{\cal{I}})$ for a knowledge base in $\dlcm_p$ consists of a set of objects $\Delta_C^{\cal{I}}$, a set of datatype values $\Delta_T^{\cal{I}}$, and a function
	$\cdot^{\cal{I}}$ satisfying the constraints shown in Table \ref{tabPCPint}. 
	It is said that $\cal{I}$ {\em satisfies} the assertion $C\sqsubseteq D$ iff $C^{\cal{I}}\subseteq D^{\cal{I}}$; and it {\em satisfies} the assertion $\mathsf{id}\, C\, a$ iff there exists $T$ such that
	$C^{\cal{I}}\subseteq (\exists a.T\sqcap \leq 1 a)^{\cal{I}}$ (mandatory 1) and for all $v\in T^{\cal{I}}$ it holds that 
	$\#\{c|c\in \Delta_C^{\cal{I}}\wedge (c,v)\in a^{\cal{I}}\}\leq 1$ (inverse functional).
\end{definition}

\begin{table}
	\centering
	\caption{Semantics of {\sl \dlcm$_p$}.}\label{tabPCPint}
	\begin{tabular}{c}
		$\top^{\cal{I}} \subseteq  \Delta_C ^{\cal{I}}$ \\ 
		$A^{\cal{I}}  \subseteq \top^{\cal{I}}$ \\  
		$\top_2^{\cal{I}} = \top^{\cal{I}}\times\top^{\cal{I}} $ \\ 
		$P^{\cal{I}}  \subseteq  \top_2^{\cal{I}}$  \\ 
		$T^{\cal{I}}  \subseteq  \Delta_T ^{\cal{I}}$ \\ 
		$a^{\cal{I}} \subseteq \top^{\cal{I}}\times \Delta^{\cal{I}}_T$   \\ 
		$(C \sqcap D)^{\cal{I}} = C^{\cal{I}} \cap D^{\cal{I}}$ \\
		$  (\leq k[i] R)^{\cal{I}}  =  \{ c \in \Delta_C ^{\cal{I}} |  \# \{  (d_1,d_2) \in R^{\cal{I}}. d_i=c\} \leq k \}$ \\
		$  (\geq k[i] R)^{\cal{I}}  =  \{ c \in \Delta_C ^{\cal{I}} |  \# \{d_1,d_2) \in R^{\cal{I}}. d_i=c\} \geq k \}$  \\
		$  (\exists a.T)^{\cal{I}}  =  \{ c \in \Delta_C ^{\cal{I}} | \exists b\in T^{\cal{I}}. (c,b) \in a^{\cal{I}} \} $ \\	
$  (\forall a.T)^{\cal{I}}  =  \{ c \in \Delta_C ^{\cal{I}} | \forall v\in \Delta^{\cal{I}}_T. (c,v) \in a^{\cal{I}}\rightarrow v\in T^{\cal{I}} \} $ \\ 			
$ (\leq 1\, a)^{\cal{I}}  =  \{ c \in \Delta_C^{\cal{I}} | \# \{ (c,v) \in a^{\cal{I}} \} \leq 1 \}$ \\
  $(i:C)^{\cal{I}} = \{ (d_1,d_2)\in \top_2^{\cal{I}}|d_i\in C^{\cal{I}}\}$ 
	\end{tabular}
\end{table}

In total, all the entities in the core profile sum up to 87.57\% of the entities in all
the analysed models, covering 91,88\% of UML models, 73.29\% of ORM models,
and 94.64\% of EE/EER models. 
Conversely, the following have been excluded from the core despite the feature overlap, due to their low incidence in the model set: 
Role (DL role component) and Relationship (DL role) Subsumption, and Completeness and Disjointness constraints. 
This means that it is not
possible to express union and disjointness of concepts in a $\dlcm_p$ knowledge base obtained by formalising a conceptual model. Clearly they can be expressed by combinations
of the constructors in $\dlcm_p$, but this is not possible if we follow the previous construction rules.
Since completeness and disjointness constraints are not present, reasoning in this core profile is quite simple.

This logic $\dlcm_p$ can be directly embedded into $\cal{DLR}$ (attributes are treated as binary relationships, and identification constraint over attributes can represented as in \cite{Calvanese01}) 
which gives ExpTime worst case complexity for satisfiability and logical implication.
A lower complexity would be expected due to the limitations in the expressivenes. For example, completeness and
disjointness constraints are not present, and negation cannot be directly expressed.
It is possible to code negation only with cardinality constraints \cite[chapter 3]{Baader08}, but
then we need to reify each negated concept as a new idempotent role, which is not possible to get from the $\dlcm_p$ rules. Another
form of getting contradiction in this context is by setting several cardinality
constraints on the same relationship participation, which is also disallowed in the rules. In any case, the main reasoning problems on the conceptual model
are only class subsumption and class equivalence on the given set of axioms. 

In spite of all these limitations, no simpler positionalist DL has been introduced. 
In order to get lower complexity bounds we need to translate a $\dlcm_p$ TBox to a standard (non-positionalist) logic, like $\dlcm_s$ below.

\begin{algorithm}[h]
	\caption{{\sl Positionalist Core to Standard Core}}
	\label{alg:eer}
	\begin{algorithmic}
		\STATE $P$ an atomic binary relationship; $D_P$ domain of $P$; $R_P$ range of $P$
		\IF{$D_P \neq R_P$ }
		\STATE Rename $P$ to two `directional' readings, $Pe_1$ and $Pe_2$
		\STATE Make $Pe_1$ and $Pe_2$ a DL relation (role) 
\STATE Type the relations with $\exists Pe_1  \sqsubseteq \forall D_P$ and $\exists  Pe_1^- \sqsubseteq R_P$
		\STATE Declare inverses with  $Pe_1 \equiv Pe_2^-$
		\ELSE \IF{$D_P = R_P$}
		\IF{$i=1,2$ is named} 
		\STATE $Pe_i \leftarrow i$		
		\ELSE 
		\STATE $Pe_i \leftarrow $ user-added label or auto generated label
		\ENDIF	
		\STATE Make $Pe_i$ a DL relation (role)	
		\STATE Type one $Pe_i$, i.e., $\exists Pe_i  \sqsubseteq D_P$ and $\exists Pe_i^- \subseteq R_P$
		\STATE Declare inverses with $Pe_i \equiv Pe_2^-$
		\ENDIF
		\ENDIF
	\end{algorithmic}
\end{algorithm}


\subsubsection{Standard Core Profile}
\label{sec:score}

Considering formalisation choices such as the positionalism of the relationships \cite{Keet09orm,Leo08} and whether to use 
inverses or qualified cardinality constraints, a {\em standard core profile} has been specified \cite{FK15adbis}. 
In case the original context is a positionalist language, a translation into a standard (role-less) language is required. Algorithm \ref{alg:eer} (adapted from \cite{FK15adbis}) does this work
in linear time in the number of elements of the vocabulary. The main step involves recursive binary relations that
generally do have their named relationship components vs `plain' binaries that
have only the relationship named.

\begin{definition}\label{def:dlcms}
	Given a conceptual model in any of the analysed CDMLs, we construct a {\em knowledge based} in $\dlcm_s$ by applying algorithm \ref{alg:eer} to its $\dlcm_p$ knowledge base. 
\end{definition}
Again, the algorithm is linear in the number of binary relationships in the knowledge base, not affecting complexity results when reasoning.

Once this conversion step is done, the formalisation of the standard core profile  is described as follows. It includes inverse relations to keep connected both relationships generated by reifying roles. 
Take atomic binary relations ($P$), atomic concepts ($A$), and simple attributes ($a$) as the basic elements of the core profile language $\dlcm_s$, which allows us to construct binary relations and 
arbitrary concepts according to the following syntax:
\begin{align*}
C &\longrightarrow  \top_1 \vsep A \vsep \forall R.A \vsep \exists R.A \vsep \leq k\, R \vsep \geq k\, R \vsep \forall a.T \vsep \\
    & \mbox{\hspace{8mm}}  \exists a.T \vsep \leq 1\, a.T \vsep C \sqcap D \\
R &\longrightarrow   \top_2 \vsep P \vsep P^- \\
X &\longrightarrow  C\sqsubseteq D \vsep \mathsf{id}\, C\, a
\end{align*}

\begin{definition}
	A {\em $\dlcm_s$ interpretation} for a knowledge base in $\dlcm_s$ is given by $\cal{I}= (\Delta^{\cal{I}}_C, \Delta^{\cal{I}}_T,\cdot^{\cal{I}})$,  with $\Delta ^{\cal{I}}_c$ the domain of interpretation for concepts,
	$\Delta^{\cal{I}}_T$ the domain of datatype values, and the interpretation function $\cdot ^{\cal{I}}$ satisfying the conditions in  Table \ref{tabSCPint}.
	$\cal{I}$ {\em satisfies} an axiom $X$ as in $\dlcm_p$.
\end{definition}

\begin{table}
	\centering
	\caption{Semantics of $\dlcm_s$.} \label{tabSCPint}
	\begin{tabular}{ccc}
		$\top^{\cal{I}}  \subseteq  \Delta_C ^{\cal{I}}$ \\  
		$A^{\cal{I}}  \subseteq  \top^{\cal{I}}$ \\ 
		$\top_2^{\cal{I}} = \top^{\cal{I}}\times \top^{\cal{I}}$                               \\ 
		$P^{\cal{I}}  \subseteq  \top_2^{\cal{I}}$  \\ 
		$T^{\cal{I}}  \subseteq  \Delta_T ^{\cal{I}}$ \\ 
		$a^{\cal{I}} \subseteq \top^{\cal{I}}\times \Delta^{\cal{I}}_T$ \\ 
		$(C \sqcap D)^{\cal{I}} = C^{\cal{I}} \cap D^{\cal{I}}$  \\ 
		& &   \\
		$(R^-)^{\cal{I}} = \{ (c_2,c_1) \in \Delta ^{\cal{I}}_C \times \Delta ^{\cal{I}}_C | (c_1,c_2) \in R^{\cal{I}} \}$\\
		$  (\forall R.A)^{\cal{I}}  =  \{ c_1 \in \Delta ^{\cal{I}}_C | \forall c_2. (c_1,c_2) \in R^{\cal{I}} \rightarrow c_2 \in A^{\cal{I}} \} $ \\
		$  (\exists R.A)^{\cal{I}}  =  \{ c_1 \in \Delta ^{\cal{I}}_C | \exists c_2. (c_1,c_2) \in R^{\cal{I}} \land c_2 \in A^{\cal{I}} \} $  \\
		$ (\leq k\, R)^{\cal{I}}  =  \{ c_1 \in \Delta ^{\cal{I}}_C |\, \#\{ c_2 | (c_1,c_2) \in R^{\cal{I}} \}   \leq k \}$  \\
		$ (\geq k\, R)^{\cal{I}}  =  \{ c_1 \in \Delta ^{\cal{I}}_C |\, \#\{ c_2 | (c_1,c_2) \in R^{\cal{I}} \}  \geq k \}$ \\ 
 $  (\forall a.T)^{\cal{I}}  =  \{ c \in \Delta ^{\cal{I}}_C | \forall v. (c,v) \in a^{\cal{I}} \rightarrow v \in T^{\cal{I}} \} $\\ 		
 $  (\exists a.T)^{\cal{I}}  =  \{ c \in \Delta ^{\cal{I}}_C | \exists v. (c,v) \in a^{\cal{I}} \land v \in T^{\cal{I}} \} $  \\
  $ (\leq 1\, a)^{\cal{I}}  =  \{ c \in \Delta ^{\cal{I}}_C |\, \#\{ (c,v) \in a^{\cal{I}} \} |  \leq 1 \}$  \\
	\end{tabular}
	\label{tab:sem}
\end{table}

From the perspective of reasoning over  $\dlcm_s$, this is rather simple and little can be deduced: negation cannot be directly expressed 
here either, as discussed for $\dlcm_p$. 
This leaves the main reasoning problem of  class subsumption and class equivalence here as well. At most the DL $\mathcal{ALNI}$ (called $\mathcal{PL}_1$  in \cite{donini1991tractable}) is 
expressive enough to represent this profile, since we only need $\top$, $\sqcap$, inverse roles and cardinality constraints; $\mathcal{PL}_1$ has polynomial subsumption,
but its data complexity is unknown. That said, using a similar encoding of conceptual models as given in Section~\ref{sec:poscore},
the language can be reduced further to $DL$-$Lite_{core}^{(\cal{HN})}$ which is {NLogSpace} with some restrictions on the interaction
between role inclusions and number restrictions, and the Unique name Assumption (UNA).
Observe that the $DL$-$Lite_{core}$ fragment is also  enough to include class disjointness in NLogSpace, and jumps to NP including disjoint covering \cite{Artale09thedl-lite}.


\subsubsection{UML Class diagram Profile}


The profile for UML Class Diagrams strictly extends $\dlcm_s$. It was presented extensively in 
\cite{FK15adbis} and succinctly formally specified here.
\begin{definition}
	A {\em knowledge base} in $\dlcm_{UML}$ from a given conceptual model in UML is obtained by  adding to its $\dlcm_s$ knowledge base the following formulas and axioms:
	\begin{enumerate}
		\item for each attribute cardinality $m..n$ in an attribute $a$ of datatype $T$ within an object type $A$ 
		we add the assertion  $A \sqsubseteq\, \leq n\, a.T\, \sqcap \geq m\, a.T$.
		\item for each binary relationship subsumption between relationships $R$ and $S$ we add the axiom $R\sqsubseteq S$.
	\end{enumerate}
\end{definition}
The syntax is as in $\dlcm_s$, with the additions highlighted in bold face for easy comparison:
\begin{align*}
C &\longrightarrow \top\vsep A\vsep \forall R.A\vsep \exists R.A\vsep \leq k\, R\vsep \geq k\, R\vsep \forall a.T\vsep \exists a.T\\
C &\longrightarrow \leq \textbf{\textit{k}}\, a.T\vsep\geq \textbf{\textit{k}}\, \textbf{\textit{a.T}}\vsep  C \sqcap D\\
R &\longrightarrow \top_2\vsep P\vsep P^- \\
X &\longrightarrow C\sqsubseteq D\vsep \mathbf{R\sqsubseteq S}\vsep \mathsf{id}\, C\, a
\end{align*}
With this profile, we cover $99.44\%$ of all the elements in the UML models of the test set. 
Absence of rarely used UML-specific modelling elements, such as the qualified association (relationship), completeness and disjointness among subclasses  
does limit the formal meaning of their models. 
On the positive side from a computational viewpoint, however, is that adding them to the language bumps up the complexity of reasoning over the models (to ExpTime-hardness \cite{Berardi05}); or: the advantage of their rare use is that reasoning over such limited diagrams has just becomes much more efficient than previously assumed to be needed. 
\begin{definition}
	A {\em $\dlcm_{UML}$ interpretation} for a $\dlcm_{UML}$ knowledge base  is a $\dlcm_s$ interpretation $\I$ that also satisfies $R\sqsubseteq S$ if and only if $R^{\I} \subseteq S^{\I}$, with
	$ (\leq k\, a.T)^{\cal{I}}  =  \{ c \in \Delta ^{\cal{I}}_C |\, \#\{ a \in T^{\cal{I}} | (c,a) \in a^{\cal{I}} \}   \leq k \}$ and
	$ (\geq k\, a.T)^{\cal{I}}  =  \{ c \in \Delta ^{\cal{I}}_C |\, \#\{ a \in T^{\cal{I}} | (c,a) \in a^{\cal{I}} \}   \geq k \}$. 
\end{definition}

Compared to $\dlcm_s$, role hierarchies  have to be added to the $\mathcal{ALNI}$ logic of the Core Profile, which yields the logic $\mathcal{ALNHI}$. 
To the best of our knowledge, this language has not been studied yet. 
If we adjust it a little by assuming unique names and some, from the conceptual modelling point of view, reasonable restrictions on the interaction between role inclusions and cardinality constraints, 
then the UML profile can be represented in the  known $DL\mbox{-}Lite_{core}^{\cal{\{HN\}}}$, which is NLogSpace for subsumption and $AC^0$ 
for data complexity \cite{Artale09thedl-lite}.
Also, if one wants to add attribute value constraints to this profile then reasoning over concrete domains is necessary. The interaction of inverse roles and concrete domains is known to be highly intractable, 
just adding them to $\cal{ALC}$ gives ExpTime-hard concept satisfiability \cite{DBLP:conf/ijcai/BaaderBL05}.


\subsubsection{ER and EER Profile}

The profile for ER and EER Diagrams also extends $\dlcm_s$. 
\begin{definition}
	A {\em knowledge base} in $\dlcm_{EER}$ from a given conceptual model in EER is obtained by adding to its its $\dlcm_s$ knowledge base the following formulas and axioms:
	\begin{enumerate}
		\item we include atomic ternary relationships in the basic vocabulary.
		\item for each attribute cardinality $m..n$ in an attribute $a$ of datatype $T$ within an object type $A$, 
		we add the assertion  $A \sqsubseteq\, \leq n\, a.T\, \sqcap \geq m\, a.T$.
		\item for each weak identification of object type $A$ through relationship $P$ in which it participates as the $i_3$-th component,
		we add the assertion $\mathsf{fd}\, R\, i_1,i_2\rightarrow i_3$, such that $1\leq i,i_1,i_2\leq 3$ and are all different.
		\item associative object types are formalised by the reification of the association
		as a new DL concept with two binary relationships.
		\item multi-attribute identification is formalised as a new composite
		attribute with single identification.
	\end{enumerate}
\end{definition}

This profile was presented extensively in 
\cite{FK15adbis} and is here recast in shorthand DL notation. The syntax is as in $\dlcm_s$, with the additions highlighted in bold face for easy comparison: 
\begin{align*}
C &\longrightarrow \top\vsep A\vsep \forall R.A\vsep \exists R.A\vsep \leq k\, R\vsep \geq k\, R\vsep \forall a.T\vsep \exists a.T\\
C & \longrightarrow \leq \textbf{\textit{k}}\, a.T\vsep \geq \textbf{\textit{k}}\, \textbf{\textit{a.T}} \vsep  C \sqcap D\\
R &\longrightarrow \top_n\vsep P\vsep P^- \\
X &\longrightarrow C\sqsubseteq D\vsep \mathsf{id}\, C\, a\vsep \mathbf{\mathsf{fd}\, R\, i_1,i_2\rightarrow i_3}
\end{align*}
where $n=2,3$ and all $i_j=1,2,3$ and different. 
\begin{definition}
	An {\em interpretation} $\cal{I}$ satisfies a knowledge base in $\dlcm_{EER}$ is it is a $\dlcm_s$ interpretation, and satisfies $\mathsf{fd}\, R\, i_1,i_2\rightarrow i_3$ iff 
	for all $r,s\in R^{\cal{I}}$ it holds that if $[i_1]r=[i_1]s$ and $[i_2]r=[i_2]s$ then $[i_3]r=[i_3]s$, , with
	$ (\leq k\, a.T)^{\cal{I}}  =  \{ c \in \Delta ^{\cal{I}}_C |\, \#\{ a \in T^{\cal{I}} | (c,a) \in a^{\cal{I}} \}   \leq k \}$ and
	$ (\geq k\, a.T)^{\cal{I}}  =  \{ c \in \Delta ^{\cal{I}}_C |\, \#\{ a \in T^{\cal{I}} | (c,a) \in a^{\cal{I}} \}   \geq k \}$. 
\end{definition}

This profile covers relative frequent EER modelling entities such as composite and multivalued attributes, weak object types and weak identification, 
ternary relationships, associated objet types and
multiattribute identification in addition to those of the standard core profile. This profile can capture $99.06 \%$ of all the elements in the set of (E)ER models. 
Multivalued attributes can be represented with attribute cardinality constraints, and composite attributes with
the inclusion of the union datatype derivation operator.
Each object type (entity type) in
(E)ER is assumed by default to have at least one identification constraint. In
order to represent external identification (weak object types), we can use functionality constraints
on roles as in $\dlrifd$ \cite{Calvanese01} 
and its close relative $\mathcal{DLR}^+$ \cite{Artale17} 
or in $\mathcal{CFD}$ \cite{toman2009applications}.
Ternary relationships are explicitly added to the profile. If we want to preserve the identity of these relationships in the DL semantics, then we need to restrict to logics in the $\dlr$ family.
Otherwise, it is possible to convert ternaries into concepts by reification, as described in Algorithm \ref{alg:nary}, using three traditional DL roles and therefore allowing the translation into logics such 
as $\mathcal{CFD}$. Since associative object types do not impose new static constraints on the models, they are formalised by reification of the association as a new DL concept with two binary relationships. 
Finally, multiattribute identification can be represented as a new composite attribute with single identification.

This profile presents an interesting challenge regarding existing languages. The only DL language family that has arbitrary $n$-aries and the advanced identification constraints needed
for the weak entity types is the positionalist $\dlrifd$. However, $\dlrifd$ also offers DL role components that are not strictly needed for (E)ER, so one could pursue a 
binary or $n$-ary DL without DL role components but with identification constraints, the latter being needed of itself and for reification of a $n$-ary into a binary (Algorithm~\ref{alg:nary}). The $\mathcal{CFD}$ family of languages may seem more suitable, then. 
Using Algorithm~\ref{alg:nary}'s translation, and since we do not have covering constraints in the profile, we can represent the (E)ER Profile in the description logic $DL\mbox{-}Lite_{core}^{\cal{N}}$ \cite{Artale09thedl-lite}
which has complexity NLogSpace for the satisfiability problem. This low complexity is in no small part thanks to its UNA,  
whereas most logics operate under no unique name assumption. A similar result is found in  \cite{Artale07er} for $ER_{ref}$, but it excludes composite attributes and weak object types. 

\begin{algorithm}[h]
	\caption{{\sl Equivalence-preserving n-ary into a binary conversion}}
	\label{alg:nary}
	\begin{algorithmic}
		\STATE v$D_P$: domain of $P$; $R_P$ range of $P$; $n$ set of $P$-components
		\STATE Reify $P$ into $P' \sqsubseteq \top$
		\FORALL{$i$,  $3 \geq i \geq n$} 
		\STATE $Re_i \leftarrow $ user-added label or auto generated label
		\STATE Make $Re_i$ a DL role, 
		\STATE Type $Re_i$ as  
		$\exists Re_i \sqsubseteq P'$ and $\exists Re_i^- \sqsubseteq R_P$, where $R_P$ is the player ((E)ER entity type) in $n$
		\STATE Add $P' \sqsubseteq \exists Re_i.\top$ and $P' \sqsubseteq\, \leq 1\, Re_i.\top$		
		\ENDFOR
		\STATE Add external identifier   $\top \sqsubseteq\, \leq 1\, (\sqcup_i Re_i)^{-}.P'$  
	\end{algorithmic}
\end{algorithm}


\subsubsection{ORM and ORM2 Profile}
\label{sec:ormprofile}

Unlike for the ER and EER profile, there is no good way to avoid the ORM roles (DL role components), as they are used for several constraints that have to be included. Therefore, to realise this profile, we must 
transform the ORM positionalist commitment into a standard view, as we did in Algorithm \ref{alg:eer}. This is motivated by the observation that typically fact type readings are provided, 
not user-named ORM role names, and only 9.5\% of all ORM roles in the 33 ORM diagrams in our dataset had a user-defined name, with a median of 0.
We process the fact type (relationship $P$) readings and ignore the role names following  Algorithm~\ref{alg:orm}.
$\mathcal{DLR}$'s relationship is typed, w.l.o.g. as binary and in $\mathcal{DLR}$-notation, as $P \sqsubseteq [r_c]C \sqcap [r_d]D$, with $r_c$ and $r_d$ variables for the ORM role names and $C$ and $D$ 
the participating object types. Let $read_1$ and $read_2$ be the fact type readings, then use $read_1$ to name DL role $Re_1$ and $read_2$ to name DL role $Re_2$, and type $P$ as 
$\top \sqsubseteq \forall Re_1.C \sqcap \forall Re_2.D$. This turns, e.g., a disjoint constraints between ORM roles $r_c$ of relationship $P$ and $s_c$ of $S$ into 
$Re_1 \sqsubseteq \neg Se_1$ and $Se_1\sqsubseteq \neg Re_1$.

\begin{algorithm}[h]
	\caption{{\sl ORM/2 to standard view and common core.}}
	\label{alg:orm}
	\begin{algorithmic}
		\STATE $P$ an atomic relationship
		\IF{$P$ is binary}
		\STATE Take fact type readings $F$
		\IF{there is only one fact type reading}
		\STATE $Re_1 \leftarrow F$
		\STATE Type $Re_1$ with domain and range
		\STATE Create $Re_2$ 
		\STATE Declare $Re_1$ and $Re_2$ inverses
		\ELSE
		\STATE Assign one reading to $Re_1$ and the other to $Re_2$
		\STATE Type $Re_1$ with domain and range	 accordingly
		\STATE Declare $Re_1$ and $Re_2$ inverses
		\ENDIF
		\ELSE \STATE $P$ is $n$-ary with $n>2$
		\STATE Reify $P$ into $P' \sqsubseteq \top$, like in Algorithm~\ref{alg:nary}, with for the $n$ binaries using the fact type readings as above 
		\ENDIF
	\end{algorithmic}
\end{algorithm}

The profile for ORM/2 Diagrams  was presented in 
\cite{FK15adbis}, and a more detailed version including a text-based mapping as a restricted ``$\ormcfd$''  was developed in \cite{FKT15} using \logicm~ as underlying logic, yet 
that could cover only just over 96\% of the elements in the set of ORM models, whereas this one reaches 98.69\% coverage.
\begin{definition}
	A {\em knowledge base} in $\dlcm_{ORM}$ from a given conceptual model in ORM2 is obtained by adding to its 
	$\dlcm_s$ knowledge base the following formulas and axioms:
	\begin{enumerate} 
		\item each $n$-ary relationship is reified as in Algorithm~\ref{alg:orm}.
		\item each unary role is formalised as a boolean attribute.
		\item add pairwise disjoint axioms for each pair of relationships with different arity.
		\item each subsumption between roles $R,S$ is represented by the formula $R\sqsubseteq S$.
		\item each subsumption between relationships $R,S$  is represented by the subsumption between the reified concepts, $R'\sqsubseteq S'$,  and the subsumption of each of the $n$ components of the relationships, $R_{ei} \sqsubseteq S_{ei}, 1\leq i \leq n$.
		\item each disjoint constraint between roles $R$ and $S$ is formalised as two
		inclusion axioms for roles: $R\sqsubseteq \neg S$ and $S\sqsubseteq \neg R$.
		\item each nested object type is represented by  the reified
		concept of the relationship.
		\item each value constraint is represented by a new datatype that constraint.
		\item each disjunctive mandatory constraint for object type $A$  in roles $R_i$ is 
		formalised as the inclusion axiom $A \sqsubseteq \sqcup_i \exists R_i$.
		\item each internal uniqueness constraint for roles $R_i , 1 \leq i \leq n$ over relationship
		objectified with object type $A$  is represented by  $\mathsf{id}\, A\, 1 R_1,\ldots, 1 R_n$
		\item each external uniqueness constraint between roles $R_i, 1 < i \leq n$
		not belonging to the same relationship is represented by 
		$\mathsf{id}\, A\, 1 R_1,\ldots, 1 R_n$, where $A$ 
		is the connected object type between all the
		$R_i$, if it exists, or otherwise a new object type representing the reification
		of a new $n$-ary relationship between the participating roles.
		\item each  external identification is represented as the previous one, with the 
		exception that we are now sure such $A$ exists. 
		(i.e., the mandatoryness is added cf. simple uniqueness). 
	\end{enumerate}
\end{definition}
This slightly more comprehensive language is here recast in shorthand DL notation, with the additions highlighted in bold face for easy comparison:
\begin{align*}
C &\longrightarrow \top_1\vsep A \vsep \forall R.A\vsep \exists R.A\vsep \leq k\, R\vsep \geq k\, R\vsep \forall a.T\vsep \\
    & \mbox{\hspace{8mm}} \exists a.T\vsep \leq 1\, a.T\\
C&\longrightarrow  C \sqcap D\vsep \mathbf{C\sqcup D}\\
R &\longrightarrow \top_2\vsep P\vsep P^-\vsep \mathbf{\neg R} \\
X &\longrightarrow C\sqsubseteq D\vsep \mathbf{R\sqsubseteq S}\vsep \mathsf{id}\, C\, a\vsep \mathbf{\mathsf{id}\, C\, R_1\ldots R_n}
\end{align*}
\begin{definition}
	A $\dlcm_{ORM}$ {\em interpretation} for a $\dlcm_{ORM}$ knowledge base is a $\dlcm_s$ interpretation $\cal{I}$  with
	the  constraints that $(C \sqcup D)^{\cal{I}}=C^{\cal{I}}\cup D^{\cal{I}}$, and $(\neg R)^{\cal{I}}=\top_2^{\cal{I}}\backslash R^{\cal{I}}$.
	$\cal{I}$ {\em satisfies} the assertion $R\sqsubseteq S$ iff $(R\sqsubseteq S)^{\cal{I}}=R^{\cal{I}}\subseteq S^{\cal{I}}$, and the assertion 
	$\mathsf{id}\, C\, R_1\ldots R_n$ iff 
	$C^{\cal{I}}\subseteq  \cap_i (\exists R_i\sqcap \leq 1 R_i)^{\cal{I}}$  and for all objects $d_1,\ldots,d_n\in T^{\cal{I}}$ it holds that $\#\{c|c\in C^{\cal{I}}\wedge  (c,d_i)\in R_i^{\cal{I}}, 1\leq i\leq n
	\}\leq 1$.
\end{definition}

We decided not to include any ring constraint in this profile. Although the irreflexivity constraint counts for almost half of all appearances of ring constraints, its participation
is still too low to be relevant.

The semantics, compared to $\dlcm_s$, is, like with the UML profile, extended in the interpretation for relationship subsumption. 
It also needs to be extended for the internal uniqueness, with the identification axioms for relationships. 
Concerning complexity of the ORM/2 Profile, this is not clear either. The ExpTime-complete $\dlrifd$ is the easiest fit, 
but contains more than is strictly needed: neither concept disjointness and union are needed (but only among roles), nor  its {\bf fd} for complex functional dependencies. 
The PTIME $\logicm$ \cite{TW14} may be a better candidate if we admit a similar translation as the one given in Algorithm \ref{alg:nary}, 
but giving up arbitrary number restrictions and disjunctive mandatory on ORM roles.

\subsection{Example application of the construction rules}

\begin{figure*}
	\centering
	\includegraphics[width=0.8\textwidth]{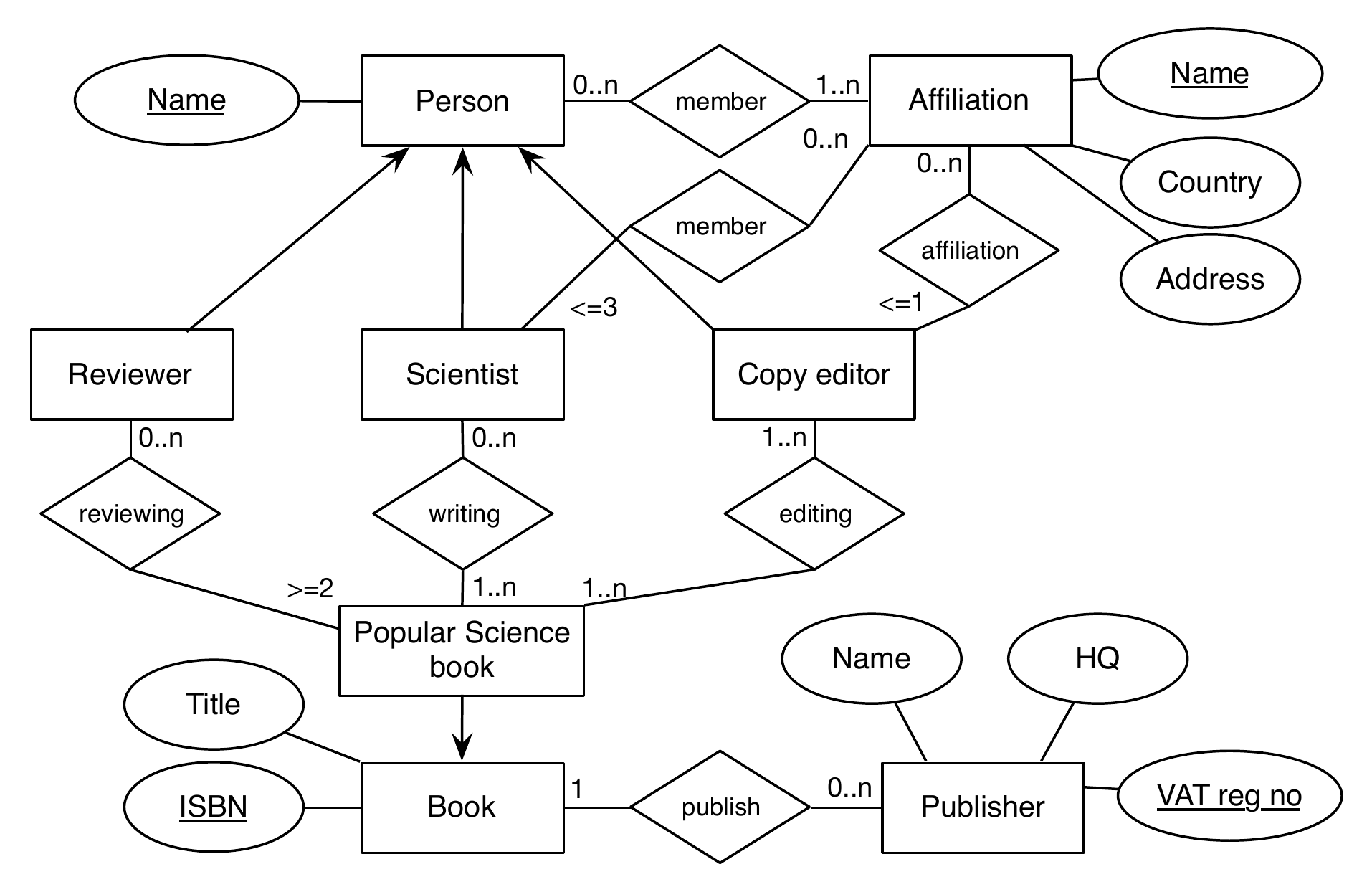}
	\includegraphics[width=0.85\textwidth]{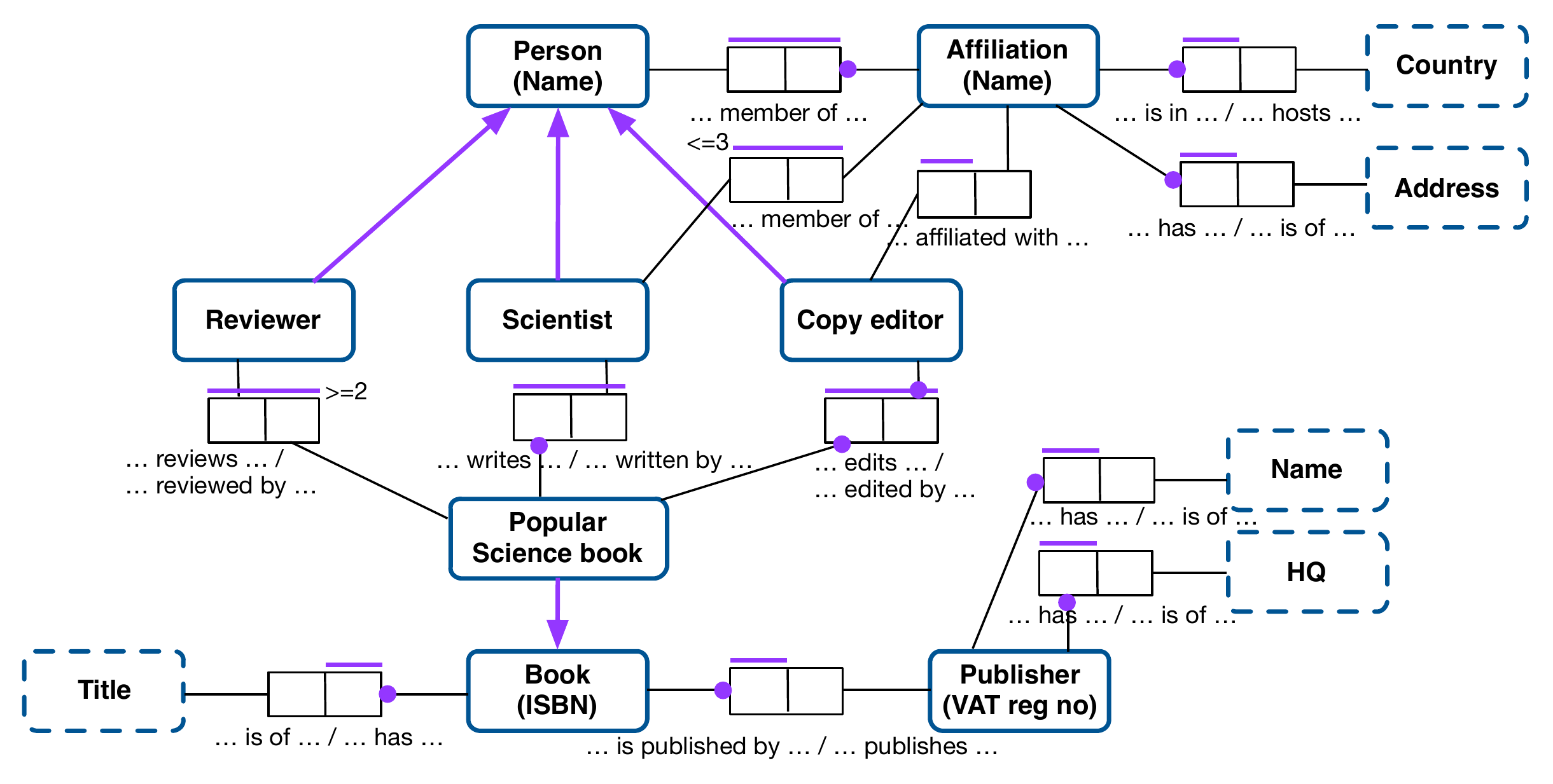}
	\caption{The sample diagram of Fig.~\ref{fig:ExUML} rendered in EER and ORM2 notation; the common $\dlcm_s$ logic-based construction is discussed in the text.}\label{fig:ExampleDiagrams}
\end{figure*}

Let us now return to the claim in the introduction about the sample UML Class Diagram in Fig.~\ref{fig:ExUML}: that it has a logical underpinning in $\dlcm_s$ and therewith also has grounded equivalents in EER and ORM notation. The equivalents in EER and ORM are shown in Fig.~\ref{fig:ExampleDiagrams}. 

The first step is to note that the $\dlcm_s$ reconstruction is obtained from $\dlcm_p$+ Algorithm~\ref{alg:eer} (by Definition~\ref{def:dlcms}). By the 
$\dlcm_p$ rules from Definition~\ref{def:dlcmp}, we obtain the set of object types (fltr) {\sf \{Person, Affiliation, ..., Publisher\}} and of data types {\sf \{Name, ..., VAT reg no\}}.
For the relationships, we need to use Algorithm~\ref{alg:eer}, which we illustrate here for the association between {\sf Person} and {\sf Affiliation}:
1)  bump up the association end names, {\sf has member} and {\sf has}, to DL roles; 2) type the relationships with: \\
$ {\tt \top \sqsubseteq \forall hasMember.Affiliation \sqcap \forall hasMember^-.Person} $\\
$ {\tt \top \sqsubseteq \forall has.Person \sqcap \forall has^-.Affiliation} $\\ 
and 3) declare inverses, 
${\tt hasMember \equiv has^-}$. 
After doing this for each association in the diagram, we continue with step 3 of Definition~\ref{def:dlcmp}, being the attributes. For instance, the {\sf Person}'s {\sf Name} we obtain  the axiom\\
$ {\tt Person\sqsubseteq \exists Name.String\, \sqcap \leq 1~ name} $\\
and likewise for the other attributes.
Step 4 takes care of the subsumptions; among others\\ 
${\tt Popular\_science\_book\sqsubseteq Book}$\\ 
is added to the $\dlcm_s$ knowledge base. Then cardinalities are processed in steps 5 and 6 (noting the algorithmic conversion from positionalist to standard view applies in this step), so that, for the membership association illustrated above, the following axioms are added to the knowledge base:  ${\tt Affiliation \sqsubseteq\, \geq 1~ has\_member}$ (mandatory participation) whereas for, say, the scientist, it will be ${\tt Scientist \sqsubseteq\, \leq 3~ has}$. Finally, any identifiers are processed, such as {\sf ISBN} for {\sf Book}, generating the addition of the $\mathsf{id}\,{\tt Book\,ISBN}$ to the $\dlcm_s$ knowledge base.

The process for the EER diagram is the same except that the name of the relationship can be used directly cf. bumping up the role names to relationship names. The reconstruction into ORM has two permutations cf. the UML one, which are covered by step 3 in Definition~\ref{def:dlcmp}, being the conversion algorithm from ORM's value types to attributes as described in \cite{FK14}, and it passes through the second {\tt else} statement of Algorithm~\ref{alg:eer} cf. the first {\tt if} statement that we used for UML when going from positionalist to standard view.

Diagram construction rules, i.e., going in the direction from the logic-based profile to a graphical notation, can follow the same process in reverse. This can be achieved automatically, except where labels have to be generated. For instance, if one were to have a scenario on an interoperability tool of ``UML diagram $\rightarrow$ $\dlcm_s$ $\rightarrow$ ORM diagram'' and one wants to have the fact type readings, they will have to be added, which a user could write herself or it could be generated by one of the extant realisation  engines for the controlled natural language\footnote{It would have rules that render, e.g., a ${\tt has\_member}$ into {\sf ... has member ...} and a ${\tt has\_member}^-$ into {\sf ... member of ...}}, similar to OWL verbalisation \cite{Safwat16} or SimpleNLG for natural language generation \cite{Gatt09}.


\section{Discussion}
\label{sec:disc}

The methodological approach proposed is expected to be of use for similar research to inform better the language design process and elucidate ontological commitments that are otherwise mostly hidden. The five profiles form an orchestrated network of compatible logics, which serve as the logic-based reconstructions of fragments of the three main CDMLs that include their most used features. In the remainder of the section, we discuss language design and computational complexity, and take a look ahead at applicability.

\paragraph{Language design}
To the best of our knowledge, there is no `cookbook process' for logic or conceptual data modelling language design. Frank's waterfall process \cite{Frank13} provided useful initial guidance for a methodological approach. 
In our experience in designing the profiles, we deemed our proposed extension with ``Ontological analysis of language features''  necessary for the conceptual modelling and knowledge representation languages setting cf. Frank's domain-specific languages.
An alternative option we considered beforehand was \cite{Karsai09}'s list of 26 guidelines, but they are too specific to DSLs to be amenable to CDML design, such as the DSL's distinction between abstract and concrete syntax and their corresponding guidelines. 

Zooming into the extra ``Ontological analysis of language features'' step, we had identified five decision points for language design with respect to ontology and several practical factors that are listed in Table~\ref{tab:consider} in Section~\ref{sec:designchoices}. 
To the best of our knowledge, it is the first attempt to scope this component of language/logic design systematically and it may spur further research into it. 
Our contribution in that regard should be seen as a starting point for a broader systematic investigation into this hitherto neglected aspect.
In making choices, we had to accommodate alternative design choices and the need to achieve high coverage. This was  addressed by designing two alternative cores---positionalist and standard view (item 2 in Section~\ref{sec:designchoices})---and, importantly, three algorithms to achieve that level of compatibility. More precisely, Algorithm~\ref{alg:eer} provides the conversion option for item 2---roles or not---in a generic way, Algorithm~\ref{alg:nary} takes case of the binaries vs $n$-aries (item 3a), and Algorithm~\ref{alg:orm} is a specific adaptation of Algorithm~\ref{alg:nary}. All profiles have data types (item 1 in Section~\ref{sec:designchoices}), for they are present in UML Class Diagrams and ORM/2, noting that it simply can be set to {\tt xsd:anyType} and thus have no influence, which is the case for (E)ER.  
Further, if the intended semantics of the aggregation association were to have been more specific in the UML standard, it would have merited inclusion in its profile (item 3b in Section~\ref{sec:designchoices}), with then the onus on the DL community to find a way to add it as a primitive to a DL. If included, it would likely also be possible to design a conversion algorithm between the new primitive and a plain DL role with properties. Regarding adding more types of entity types to the language (item 3c), like sortal and phase: 
the one proposal \cite{Braga10,Guizzardi05} is not in widespread use and therewith did not meet the evidence-based threshold for inclusion. 
It is also not clear how to represent in a decidable language such notions that are essentially based on OntoClean \cite{Guarino09oc} that requires modality and higher-order predicates, nor how an equivalence-preserving algorithm would look like, if possible at all.

\paragraph{Complexity considerations for the profiles}

Traditionally, the DL research community has strived for
identifying more and more expressive DLs for which reasoning is
still decidable. The introduced profiles show that high expressivity is not necessary for representing most of the semantics of conceptual models, independently 
of the chosen modelling language. 
They thus are `lean', evidence-based, profiles that, while not covering all corners of modelling issues, do have those features that are used most in practice. 
We summarise the complexity of each profile by immersion into a DL language in Table~\ref{fig:prof}. 
The ``Approximate DL'' column
is not an exact match for each profile, and often involves some extra assumptions that explains the different complexities.
Low complexities are achievable by the standard profiles (i.e., those that give up on positionalism), due to the existence of a more accurate matching logic.
Recall that $\dlcm_s$ is included in $\dlcm_{UML}$, $\dlcm_{EER}$, and $\dlcm_{ORM}$.  
The biggest gap between the profiles and the matching DLs is given in $\dlcm_p$ showing more
work on positionalist DLs is necessary, especially with respect to reasoning algorithms.  

\begin{table*}[t]
	\begin{center}
		\caption{Profile comparison on language and complexity; ``Approx. DL'': the 
			existing 
			DL nearest to the profile defined.}\label{fig:prof}
		\begin{tabular}{|p{1.2cm}|p{9.3cm}|c|p{2cm}|} \hline 
			
			\textbf{Profile} & \textbf{Main features} & \textbf{Approx. DL} & \textbf{Subsumption complexity} \\ \hline \hline
			
			$\dlcm_p$ & positionalist, binary relationships, identifiers, cardinality constraints, attribute typing, mandatory attribute and its functionality                  & $\cal{DLR}$ & ExpTime \\ \hline
			
			$\dlcm_s$ & standard view, binary relationships, inverses &           $\cal{ALNI}$ & P \\ \hline
			
			$\dlcm_{UML}$ & relationship subsumption, attribute cardinality & $\dllite_{core}^{\cal{HN}}$ & NLogSpace \\ \hline
			
			$\dlcm_{EER}$ & ternary relationships, attribute cardinality, external keys & $\dllite_{core}^{N}$ & NLogSpace \\ \cline{3-4}
			&  & $\cal{CFD}$  & P  \\ \hline
			
			$\dlcm_{ORM}$ & entity type disjunction, relationships complement, relationship  & $\dlr_{ifd}$ & ExpTime \\ \cline{3-4}
			& subsumption, complex identifiers (`multi attribute keys') &  $\logicm$ & P \\ \hline 
		\end{tabular}
	\end{center}
\end{table*}

An outstanding issue is whether object types in the diagrams are by default disjoint when not in a hierarchy, or not. Some research are convinced they are, and some are not; most formalisations and tools do not include it. 
Because of the lack of agreement, we have not included it. Note further that if this assumption were to be added, i.e., full negation in the profiles, it would affect the computational complexity of the profiles negatively. 

It is also interesting to analyse at which point increasing expressiveness by adding new features to the language is worthwhile from the point of view of the modeller. 
If the feature is present, at least one modeller will use it, though mostly only occasionally. It is not clear if 
this is due to them being corner cases, a lack of experience on representing advanced constraints by modellers, tooling, or another reason.
On the other hand UML's aggregation as `extra' feature cf. (E)ER's and ORM/2's plain relationships {\em is} being used disproportionally more often than part-whole relations in (E)ER and ORM/2. It remains to be investigated why exactly this is the case.

\paragraph{Toward applicability}

The presented profiles may be applied as the back-end of CASE tools using the compatible profiles as unifying logics and orchestration of corresponding optimised reasoners for, say, 
Ontology-Based Data Access 
such that it focusses on the perceived language needs of the modellers (cf. the logic and technology, as in, e.g., \cite{Calvanese17,Ozcep15}), whilst still keeping it tractable. 
The current conceptual modelling tools that have a logic back-end are still sparse \cite{Braga10,Farre13,Fillottrani11,Braun17}, 
and allow a modeller to model in only one language, rather than being allowed to switch between language families.

Using the common core for model interoperability by mapping each graphical element into a construct in $\dlcm_s$ is an option. 
However, one also would want to be precise and therefore use more language features than those in the common core, and  when linking models, `mismatch' links would still need to be managed, 
and wrong ones discarded. 
To solve this, an interoperability approach with equivalence, transformation, and approximation rules that is guided by the metamodel is possible \cite{FK14,KKFC16}. 
There, one can have two models with an intermodel assertion; e.g., between a UML association and an ORM fact type. The entities are first classified/mapped into entities of the metamodel, any relevant rules are executed, and out comes the result, being either a valid or an invalid link. The `any relevant rules are executed' is coordinated by the metamodel; e.g., the metamodel states that each Relationship has to have two or more Roles, which, in turn, have to have attached to it either an Object Type or Value Type, so those mapping and transformation rules are called as well during the checking of the link. 
The MIST EER tool \cite{Dimitrieski15} has a similar goal, though currently it supports only EER and its translation to SQL and therewith is complementary to our work presented here.

The formal foundation presented here would enable such an interface were either multiple graphical rendering in different modelling language families could be generated, or 
link models represented in different languages in a system integration scenario.


\section{Conclusions}
\label{sec:concl}

A systematic logic design process was proposed that generalises and extends the DSL design process to be more broadly applicable by incorporating an ontological analysis of language features in the process. This first compilation of ontological commitments embedded in a logic design process includes, among others, the ontology of relations, the conceptual vs design features trade-off, and 3-dimensionalist vs. 4-dimensionalist commitments.

Based on this extended process with explicit ontological distinctions and the evidence of the prevalence of the features in the models, different characteristic profiles for the three conceptual data modelling language families were specified into a suitable Description Logic, which also brought with it insights into their computational complexity. 
The common core profile is of relatively low computational complexity, being in the tractable $\mathcal{ALNI}$. Without the negation, hardly any inconsistencies can be derived with the profiles, with as flip side that it is promising for scalable runtime use of conceptual data models.

We are looking into several avenues for future work, including ongoing tool development and more precise complexity results for the profiles so that it would allow special, conceptual data model-optimised, reasoners.

%
%
%
%
%

\bibliographystyle{spmpsci}
\bibliography{dl16-shortnew}  

\begin{thebibliography}{100}
\providecommand{\url}[1]{{#1}}
\providecommand{\urlprefix}{URL }
\expandafter\ifx\csname urlstyle\endcsname\relax
  \providecommand{\doi}[1]{DOI~\discretionary{}{}{}#1}\else
  \providecommand{\doi}{DOI~\discretionary{}{}{}\begingroup
  \urlstyle{rm}\Url}\fi

\bibitem{tonesd27}
Alberts, R., Calvanese, D., Giacomo, G.D., Gerber, A., Horridge, M., Kaplunova,
  A., Keet, C.M., Lembo, D., Lenzerini, M., Milicic, M., M\"oller, R.,
  Rodr\'iguez-Muro, M., Rosati, R., Sattler, U., Suntisrivaraporn, B.,
  Stefanoni, G., Turhan, A.Y., Wandelt, S., Wessel, M.: Analysis of test
  results on usage scenarios.
\newblock Deliverable {TONES-D27} v1.0, TONES Project (2008)

\bibitem{Artale07er}
Artale, A., Calvanese, D., Kontchakov, R., Ryzhikov, V., Zakharyaschev, M.:
  Reasoning over extended {ER} models.
\newblock In: C.~Parent, K.D. Schewe, V.C. Storey, B.~Thalheim (eds.)
  Proceedings of the 26th International Conference on Conceptual Modeling
  (ER'07), \emph{LNCS}, vol. 4801, pp. 277--292. Springer (2007).
\newblock Auckland, New Zealand, November 5-9, 2007

\bibitem{Artale09thedl-lite}
Artale, A., Calvanese, D., Kontchakov, R., Zakharyaschev, M.: {The {DL-Lite}
  family and relations}.
\newblock Journal of Artificial Intelligence Research \textbf{36}, 1--69 (2009)

\bibitem{Artale17}
Artale, A., Franconi, E., Pe{\~{n}}aloza, R., Sportelli, F.: A decidable very
  expressive description logic for databases.
\newblock In: C.~d'Amato, M.~Fernandez, V.~Tamma, F.~Lecue,
  P.~Cudr{\'e}-Mauroux, J.~Sequeda, C.~Lange, J.~Heflin (eds.) The Semantic Web
  -- ISWC 2017: 16th International Semantic Web Conference, \emph{LNCS}, vol.
  10587, pp. 37--52. Springer, Cham (2017).
\newblock 21--25 October 2017, Vienna, Austria

\bibitem{Artale07a}
Artale, A., Parent, C., Spaccapietra, S.: Evolving objects in temporal
  information systems.
\newblock Annals of Mathematics and Artificial Intelligence \textbf{50}(1-2),
  5--38 (2007)

\bibitem{Atzeni08}
Atzeni, P., Cappellari, P., Torlone, R., Bernstein, P.A., Gianforme, G.:
  Model-independent schema translation.
\newblock VLDB Journal \textbf{17}(6), 1347--1370 (2008)

\bibitem{DBLP:conf/ijcai/BaaderBL05}
Baader, F., Brandt, S., Lutz, C.: Pushing the {EL} envelope.
\newblock In: L.P. Kaelbling, A.~Saffiotti (eds.) IJCAI-05, Proceedings of the
  Nineteenth International Joint Conference on Artificial Intelligence,
  Edinburgh, Scotland, UK, July 30 - August 5, 2005, pp. 364--369. Professional
  Book Center (2005)

\bibitem{Baader08}
Baader, F., Calvanese, D., McGuinness, D.L., Nardi, D., Patel-Schneider, P.F.
  (eds.): The Description Logics Handbook -- Theory and Applications, 2 edn.
\newblock Cambridge University Press (2008)

\bibitem{Batsakis17}
Batsakis, S., Petrakis, E., Tachmazidis, I., Antoniou, G.: Temporal
  representation and reasoning in {OWL} 2.
\newblock Semantic Web Journal \textbf{8}(6), 981--1000 (2017)

\bibitem{Berardi05}
Berardi, D., Calvanese, D., De~Giacomo, G.: Reasoning on {U}{M}{L} class
  diagrams.
\newblock Artificial Intelligence \textbf{168}(1-2), 70--118 (2005)

\bibitem{Bloesch97}
Bloesch, A.C., Halpin, T.A.: Conceptual {Q}ueries using {C}on{Q}uer-{I}{I}.
\newblock In: Proceedings of ER'97: 16th International Conference on Conceptual
  Modeling, \emph{LNCS}, vol. 1331, pp. 113--126. Springer (1997)

\bibitem{Boyd05}
Boyd, M., McBrien, P.: Comparing and transforming between data models via an
  intermediate hypergraph data model.
\newblock Journal on Data Semantics \textbf{IV}, 69--109 (2005)

\bibitem{Braga10}
Braga, B.F.B., Almeida, J.P.A., Guizzardi, G., Benevides, A.B.: Transforming
  {OntoUML} into {A}lloy: towards conceptual model validation using a
  lightweight formal methods.
\newblock Innovations in Systems and Software Engineering \textbf{6}(1-2),
  55--63 (2010)

\bibitem{BraunGCF16}
Braun, G.A., Gimenez, C., Cecchi, L.A., Fillottrani, P.R.: Towards a
  visualisation process for ontology-based conceptual modelling.
\newblock In: Proceedings of the {IX} {ONTOBRAS} Brazilian Ontology Research
  Seminar, Curitiba, Brazil, October 3rd, 2016., pp. 107--118 (2016).
\newblock \urlprefix\url{http://ceur-ws.org/Vol-1862/paper-09.pdf}

\bibitem{Braun17}
Braun, G.A., Gim{\'{e}}nez, C., Fillottrani, P.R., Cecchi, L.A.: Towards
  conceptual modelling interoperability in a web tool for ontology engineering.
\newblock In: Proceedings of the 3rd Argentine Symposium on Ontologies and
  their Applications co-located with 46 Jornadas Argentinas de
  Inform{\'{a}}tica (46JAIIO), pp. 25--38 (2017)

\bibitem{Buitelaar14}
Buitelaar, P., Cimiano, P. (eds.): Towards the Multilingual Semantic Web:
  Principles, Methods and Applications.
\newblock Springer (2014)

\bibitem{Cabot08}
Cabot, J., Claris\'o, R., Riera, D.: Verification of {UML}/{OCL} class diagrams
  using constraint programming.
\newblock In: Model Driven Engineering, Verification, and Validation:
  Integrating Verification and Validation in MDE (MoDeVVA 2008) (2008)

\bibitem{Cadoli07}
Cadoli, M., Calvanese, D., De~Giacomo, G., Mancini, T.: Finite model reasoning
  on {U}{M}{L} class diagrams via constraint programming.
\newblock In: Proc. of AI*IA 2007, \emph{LNAI}, vol. 4733, pp. 36--47. Springer
  (2007)

\bibitem{Calvanese17}
Calvanese, D., Cogrel, B., Komla-Ebri, S., Kontchakov, R., Lanti, D., Rezk, M.,
  Rodriguez-Muro, M., Xiao, G.: Ontop: Answering {SPARQL} queries over
  relational databases.
\newblock Semantic Web Journal \textbf{8}(3), 471--487 (2017)

\bibitem{Calvanese98}
Calvanese, D., De~Giacomo, G., Lenzerini, M.: On the decidability of query
  containment under constraints.
\newblock In: Proc. of the 17th ACM SIGACT SIGMOD SIGART Sym. on Principles of
  Database Systems (PODS'98), pp. 149--158 (1998)

\bibitem{Calvanese99mu}
Calvanese, D., De~Giacomo, G., Lenzerini, M.: Reasoning in expressive
  description logics with fixpoints based on automata on infinite trees.
\newblock In: Proc. of the 16th Int. Joint Conf. on Artificial Intelligence
  (IJCAI'99), pp. 84--89 (1999)

\bibitem{Calvanese01}
Calvanese, D., De~Giacomo, G., Lenzerini, M.: Identification constraints and
  functional dependencies in description logics.
\newblock In: B.~Nebel (ed.) Proc. of the 17th Int. Joint Conf. on Artificial
  Intelligence (IJCAI 2001), pp. 155--160. Morgan Kaufmann (2001).
\newblock Seattle, Washington, USA, August 4-10, 2001

\bibitem{Calvanese07}
Calvanese, D., Giacomo, G.D., Lembo, D., Lenzerini, M., Rosati, R.: Tractable
  reasoning and efficient query answering in description logics: The
  {D}{L}-{L}ite family.
\newblock Journal of Automated Reasoning \textbf{39}(3), 385--429 (2007)

\bibitem{CKNRS10}
Calvanese, D., Keet, C.M., Nutt, W., Rodr\'iguez-Muro, M., Stefanoni, G.:
  Web-based graphical querying of databases through an ontology: the {WONDER}
  system.
\newblock In: S.Y. Shin, S.~Ossowski, M.~Schumacher, M.J. Palakal, C.C. Hung
  (eds.) Proceedings of ACM Symposium on Applied Computing (ACM SAC'10), pp.
  1389--1396. ACM (2010).
\newblock March 22-26 2010, Sierre, Switzerland

\bibitem{Calvanese99}
Calvanese, D., Lenzerini, M., Nardi, D.: Unifying class-based representation
  formalisms.
\newblock Journal of Artificial Intelligence Research \textbf{11}, 199--240
  (1999)

\bibitem{Calvanese16}
Calvanese, D., Liuzzo, P., Mosca, A., Remesal, J., Rezk, M., Rull, G.:
  Ontology-based data integration in epnet: Production and distribution of food
  during the roman empire.
\newblock Engineering Applications of Artificial Intelligence \textbf{51},
  212--229 (2016)

\bibitem{Chen76}
Chen, P.P.: The entity-relationship model---toward a unified view of data.
\newblock ACM Transactions on Database Systems \textbf{1}(1), 9--36 (1976)

\bibitem{Dimitrieski15}
Dimitrieski, V., Celikovic, M., Aleksic, S., Risti, S., Alargt, A., Lukovic,
  I.: Concepts and evaluation of the extended entity-relationship approach to
  database design in a multi-paradigm information system modeling tool.
\newblock Computer Languages, Systems \& Structures \textbf{44}(Part C), 299 --
  318 (2015)

\bibitem{donini1991tractable}
Donini, F., Lenzerini, M., Nardi, D., Nutt, W.: Tractable concept languages.
\newblock In: Proc.of IJCAI'91, vol.~91, pp. 458--463 (1991)

\bibitem{Donnelly10}
Donnelly, K.A.M.: A short communication - meta data and semantics the industry
  interface: what does the food industry think are necessary elements for
  exchange?
\newblock In: Metadata and Semantic Research: 4th International Conference,
  MTSR 2010 (2010)

\bibitem{Eiter17}
Eiter, T., Parreira, J.X., Schneider, P.: Spatial ontology-mediated query
  answering over mobility streams.
\newblock In: E.~Blomqvist, et~al. (eds.) Proceedings of the 13th Extended
  Semantic Web Conference (ESWC'17), \emph{LNCS}, vol. 10249, pp. 219--237.
  Springer (2017).
\newblock 30 May - 1 June 2017, Portoroz, Slovenia

\bibitem{Farre13}
Farr\'e, C., Queralt, A., Rull, G., Teniente, E., Urp\'i, T.: Automated
  reasoning on {UML} conceptual schemas with derived information and queries.
\newblock Information and Software Technology \textbf{55}(9), 1529 -- 1550
  (2013)

\bibitem{Fillottrani11}
Fillottrani, P.R., Franconi, E., Tessaris, S.: The {ICOM} 3.0 intelligent
  conceptual modelling tool and methodology.
\newblock Semantic Web Journal \textbf{3}(3), 293--306 (2012)

\bibitem{FK14}
Fillottrani, P.R., Keet, C.M.: Conceptual model interoperability: a
  metamodel-driven approach.
\newblock In: A.~Bikakis, et~al. (eds.) Proceedings of the 8th International
  Web Rule Symposium (RuleML'14), \emph{LNCS}, vol. 8620, pp. 52--66. Springer
  (2014).
\newblock August 18-20, 2014, Prague, Czech Republic

\bibitem{FK14tr}
Fillottrani, P.R., Keet, C.M.: {KF} metamodel formalization.
\newblock Technical Report 1078634 (2014).
\newblock Arxiv.org, 21p.

\bibitem{FK15adbis}
Fillottrani, P.R., Keet, C.M.: Evidence-based languages for conceptual data
  modelling profiles.
\newblock In: T.~Morzy, et~al. (eds.) 19th Conference on Advances in Databases
  and Information Systems (ADBIS'15), \emph{LNCS}, vol. 9282, pp. 215--229.
  Springer (2015).
\newblock 8-11 Sept, 2015, Poitiers, France

\bibitem{FK16dl}
Fillottrani, P.R., Keet, C.M.: A design for coordinated and logics-mediated
  conceptual modelling.
\newblock In: R.~Pe\~naloza, M.~Lenzerini (eds.) Proceedings of the 29th
  International Workshop on Description Logics (DL'16), \emph{CEUR-WS}, vol.
  1577 (2016).
\newblock 22-25 April, 2016, Cape Town, South Africa

\bibitem{FKT15}
Fillottrani, P.R., Keet, C.M., Toman, D.: Polynomial encoding of orm conceptual
  models in $\mathcal{CFDI}_{nc}^{\forall -}$.
\newblock In: D.~Calvanese, B.~Konev (eds.) Proceedings of the 28th
  International Workshop on Description Logics (DL'15), \emph{CEUR-WS}, vol.
  1350, pp. 401--414 (2015).
\newblock 7-10 June 2015, Athens, Greece

\bibitem{Fine00}
Fine, K.: Neutral relations.
\newblock The Philosophical Review \textbf{109}(1), 1--33 (2000)

\bibitem{Franconi12}
Franconi, E., Mosca, A., Solomakhin, D.: The formalisation of {ORM2} and its
  encoding in {OWL2}.
\newblock KRDB Research Centre Technical Report KRDB12-2, Faculty of Computer
  Science, Free University of Bozen-Bolzano, Italy (2012)

\bibitem{Frank13}
Frank, U.: Domain-specific modeling languages - requirements analysis and
  design guidelines.
\newblock In: I.~Reinhartz-Berger, A.~Sturm, T.~Clark, J.~Bettin, S.~Cohen
  (eds.) Domain Engineering: Product Lines, Conceptual Models, and Languages,
  pp. 133--157. Springer (2013)

\bibitem{Gatt09}
Gatt, A., Reiter, E.: Simplenlg: A realisation engine for practical
  applications.
\newblock In: E.~Krahmer, M.~Theune (eds.) Proceedings of the 12th European
  Workshop on Natural Language Generation (ENLG'09), pp. 90--93. ACL (2009).
\newblock March 30-31, 2009, Athens, Greece

\bibitem{Guarino09a}
Guarino, N.: The ontological level: Revisiting 30 years of knowledge
  representation.
\newblock In: A.~Borgida, et~al. (eds.) Mylopoulos Festschrift, \emph{LNCS},
  vol. 5600, pp. 52--67. Springer (2009)

\bibitem{Guarino06}
Guarino, N., Guizzardi, G.: In the defense of ontological foundations for
  conceptual modeling.
\newblock Scandinavian Journal of Information Systems \textbf{18}(1), (debate
  forum, 9p) (2006)

\bibitem{Guarino09oc}
Guarino, N., Welty, C.: An overview of {OntoClean}.
\newblock In: S.~Staab, R.~Studer (eds.) Handbook on Ontologies, pp. 201--220.
  Springer Verlag (2009)

\bibitem{Guizzardi05}
Guizzardi, G.: Ontological foundations for structural conceptual models.
\newblock Phd thesis, University of Twente, The Netherlands. Telematica
  Instituut Fundamental Research Series No. 15 (2005)

\bibitem{Guizzardi10}
Guizzardi, G.: On the representation of quantities and their parts in
  conceptual modeling.
\newblock In: Proceedings of 6th International conference on Formal Ontology in
  Information Systems (FOIS'10). IOS Press (2010).
\newblock 11th-14th May 2010, Toronto, Canada

\bibitem{Guizzardi08}
Guizzardi, G., Wagner, G.: What's in a relationship: An ontological analysis.
\newblock In: Q.~Li, S.~Spaccapietra, E.S.K. Yu, A.~Oliv\'e (eds.) Proceedings
  of the 27th International Conference on Conceptual Modeling (ER'08),
  \emph{LNCS}, vol. 5231, pp. 83--97. Springer (2008).
\newblock {Barcelona}, Spain, October 20-24, 2008

\bibitem{guizzardi2010}
Guizzardi, G., Wagner, G.: Using the unified foundational ontology ({UFO}) as a
  foundation for general conceptual modeling languages.
\newblock In: Theory and Applications of Ontology: Computer Applications, pp.
  175--196. Springer (2010)

\bibitem{Halpin89}
Halpin, T.: A logical analysis of information systems: static aspects of the
  data-oriented perspective.
\newblock Ph.D. thesis, University of Queensland, Australia (1989)

\bibitem{Halpin01}
Halpin, T.: Information Modeling and Relational Databases.
\newblock San Francisco: Morgan Kaufmann Publishers (2001)

\bibitem{Halpin08}
Halpin, T., Morgan, T.: Information modeling and relational databases, 2nd edn.
\newblock Morgan Kaufmann (2008)

\bibitem{Halpin04}
Halpin, T.A.: Advanced Topics in Database Research, vol.~3, chap. Comparing
  Metamodels for ER, ORM and UML Data Models, pp. 23--44.
\newblock Idea Publishing Group, Hershey PA, USA (2004)

\bibitem{Hofstede98}
Hofstede, A.H.M.t., Proper, H.A.: How to formalize it? formalization principles
  for information systems development methods.
\newblock Information and Software Technology \textbf{40}(10), 519--540 (1998)

\bibitem{Horrocks06}
Horrocks, I., Kutz, O., Sattler, U.: The even more irresistible
  $\mathcal{SROIQ}$.
\newblock Proceedings of KR-2006 pp. 452--457 (2006)

\bibitem{Jahangard11}
Jahangard~Rafsanjani, A., Mirian-Hosseinabadi, S.H.: {A Z Approach to
  Formalization and Validation of ORM Models}.
\newblock In: E.~Ariwa, E.~El-Qawasmeh (eds.) Digital Enterprise and
  Information Systems, \emph{CCIS}, vol. 194, pp. 513--526. Springer (2011)

\bibitem{KalayciXRKC18}
Kalayci, E.G., Xiao, G., Ryzhikov, V., Kalayci, T.E., Calvanese, D.:
  Ontop-temporal: {A} tool for ontology-based query answering over temporal
  data.
\newblock In: Proceedings of the 27th {ACM} International Conference on
  Information and Knowledge Management, {CIKM} 2018, Torino, Italy, October
  22-26, 2018, pp. 1927--1930 (2018)

\bibitem{Kaneiwa06}
Kaneiwa, K., Satoh, K.: Consistency checking algorithms for restricted
  {U}{M}{L} class diagrams.
\newblock In: Proceedings of the 4th International Symposium on Foundations of
  Information and Knowledge Systems (FoIKS'06). Springer Verlag (2006)

\bibitem{Karsai09}
Karsai, G., Krahn, H., Pinkernell, C., Rumpe, B., Schindler, M., V\"olkel, S.:
  Design guidelines for {Domain Specific Languages}.
\newblock In: Proceedings of the 9th OOPSLA Workshop on Domain-Specific
  Modeling (DSM'09) (2009).
\newblock {O}rlando, Florida, USA, October 2009

\bibitem{Keet09arxiv}
Keet, C.M.: Mapping the {O}bject-{R}ole {M}odeling language {O}{R}{M}2 into
  {D}escription {L}ogic language $\mathcal{DLR}_{ifd}$.
\newblock Tech. Rep. 0702089v2, KRDB Research Centre, Free University of
  Bozen-Bolzano, Italy (2009).
\newblock ArXiv:cs.LO/0702089v2

\bibitem{Keet09orm}
Keet, C.M.: Positionalism of relations and its consequences for fact-oriented
  modelling.
\newblock In: R.~Meersman, P.~Herrero, D.~T. (eds.) OTM Workshops,
  International Workshop on Fact-Oriented Modeling (ORM'09), \emph{LNCS}, vol.
  5872, pp. 735--744. Springer (2009).
\newblock Vilamoura, Portugal, November 4-6, 2009

\bibitem{Keet12odcm}
Keet, C.M.: Ontology-driven formal conceptual data modeling for biological data
  analysis.
\newblock In: M.~Elloumi, A.Y. Zomaya (eds.) Biological Knowledge Discovery
  Handbook: Preprocessing, Mining and Postprocessing of Biological Data,
  chap.~6, pp. 129--154. Wiley (2013)

\bibitem{Keet16stuff}
Keet, C.M.: Relating some stuff to other stuff.
\newblock In: E.~Blomqvist, P.~Ciancarini, F.~Poggi, F.~Vitali (eds.)
  Proceedings of the 20th International Conference on Knowledge Engineering and
  Knowledge Management (EKAW'16), \emph{LNAI}, vol. 10024, pp. 368--383.
  Springer (2016).
\newblock 19-23 November 2016, Bologna, Italy

\bibitem{KB17}
Keet, C.M., Berman, S.: Determining the preferred representation of temporal
  constraints in conceptual models.
\newblock In: H.~Mayr, et~al. (eds.) 36th International Conference on
  Conceptual Modeling (ER'17), \emph{LNCS}, vol. 10650, pp. 437--450. Springer
  (2017).
\newblock 6-9 Nov 2017, Valencia, Spain

\bibitem{KC16}
Keet, C.M., Chirema, T.: A model for verbalising relations with roles in
  multiple languages.
\newblock In: E.~Blomqvist, P.~Ciancarini, F.~Poggi, F.~Vitali (eds.)
  Proceedings of the 20th International Conference on Knowledge Engineering and
  Knowledge Management (EKAW'16), \emph{LNAI}, vol. 10024, pp. 384--399.
  Springer (2016).
\newblock 19-23 November 2016, Bologna, Italy

\bibitem{KFM12}
Keet, C.M., Fern\'andez-Reyes, F.C., Morales-Gonz\'alez, A.: Representing
  mereotopological relations in {OWL} ontologies with {\sc {o}nto{p}art{s}}.
\newblock In: E.~Simperl, et~al. (eds.) Proceedings of the 9th Extended
  Semantic Web Conference (ESWC'12), \emph{LNCS}, vol. 7295, pp. 240--254.
  Springer (2012).
\newblock 29-31 May 2012, Heraklion, Crete, Greece

\bibitem{KF15er}
Keet, C.M., Fillottrani, P.R.: An analysis and characterisation of publicly
  available conceptual models.
\newblock In: P.~Johannesson, M.L. Lee, S.~Liddle, A.L. Opdahl,
  O.~Pastor~L\'opez (eds.) Proceedings of the 34th International Conference on
  Conceptual Modeling (ER'15), \emph{LNCS}, vol. 9381, pp. 585--593. Springer
  (2015).
\newblock 19-22 Oct, Stockholm, Sweden

\bibitem{KF15dke}
Keet, C.M., Fillottrani, P.R.: An ontology-driven unifying metamodel of {UML
  Class Diagrams, EER, and ORM2}.
\newblock Data \& Knowledge Engineering \textbf{98}, 30--53 (2015)

\bibitem{KK13medi}
Khan, Z., Keet, C.M.: The foundational ontology library {ROMULUS}.
\newblock In: A.~Cuzzocrea, S.~Maabout (eds.) Proceedings of the 3rd
  International Conference on Model \& Data Engineering (MEDI'13), \emph{LNCS},
  vol. 8216, pp. 200--211. Springer (2013).
\newblock September 25-27, 2013, Amantea, Calabria, Italy

\bibitem{KKFC16}
Khan, Z.C., Keet, C.M., Fillottrani, P.R., Cenci, K.: Experimentally motivated
  transformations for intermodel links between conceptual models.
\newblock In: J.~Pokorn\'y, et~al. (eds.) 20th Conference on Advances in
  Databases and Information Systems (ADBIS'16), \emph{LNCS}, vol. 9809, pp.
  104--118. Springer (2016).
\newblock {A}ugust 28-31, Prague, Czech Republic

\bibitem{deKinderen15}
de~Kinderen, S., Ma, Q.: Requirements engineering for the design of conceptual
  modeling languages.
\newblock Applied Ontology \textbf{10}(1), 7--24 (2015)

\bibitem{Leo08}
Leo, J.: Modeling relations.
\newblock Journal of Philosophical Logic \textbf{37}, 353--385 (2008)

\bibitem{Leo16}
Leo, J.: Coordinate-free logic.
\newblock The Review of Symbolic Logic \textbf{9}(3), 522--555 (2016)

\bibitem{Malavolta13}
Malavolta, I., Lago, P., Muccini, H., Pelliccione, P., Tang, A.: What industry
  needs from architectural languages: A survey.
\newblock IEEE Transactions on Software Engineering \textbf{39}(6), 869--891
  (2013)

\bibitem{Moody05}
Moody, D.L.: Theoretical and practical issues in evaluating the quality of
  conceptual models: current state and future directions.
\newblock Data \& Knowledge Engineering \textbf{55}, 243--276 (2005)

\bibitem{OWL2profiles}
Motik, B., Grau, B.C., Horrocks, I., Wu, Z., Fokoue, A., Lutz, C.: {OWL} 2
  {W}eb {O}ntology {L}anguage {P}rofiles.
\newblock {W3C} recommendation, W3C (2009).
\newblock Http://www.w3.org/TR/owl2-profiles/

\bibitem{OWL2rec}
Motik, B., Patel-Schneider, P.F., Parsia, B.: {OWL} 2 web ontology language
  structural specification and functional-style syntax.
\newblock W3c recommendation, W3C (2009).
\newblock Http://www.w3.org/TR/owl2-syntax/

\bibitem{Mylopoulos90}
Mylopoulos, J., Borgida, A., Jarke, M., Koubarakis, M.: Telos: Representing
  knowledge about information systems.
\newblock ACM Transactions on Information Systems \textbf{8}(4), 325--362
  (1990)

\bibitem{Nizol14}
Nizol, M., Dillon, L.K., Stirewalt, R.E.K.: Toward tractable instantiation of
  conceptual data models using non-semantics-preserving model transformations.
\newblock In: Proceedings of the 6th International Workshop on Modeling in
  Software Engineering (MiSE'14), pp. 13--18. ACM Conference Proceedings
  (2014).
\newblock Hyderabad, India, June 02-03, 2014

\bibitem{UMLspec12}
{Object Management Group}: Superstructure specification.
\newblock Standard 2.4.1, Object Management Group (2012).
\newblock Http://www.omg.org/spec/UML/2.4.1/

\bibitem{Ozcep15}
{\"O}z{\c{c}}ep, {\"O}.L., M{\"o}ller, R., Neuenstadt, C.: Stream-query
  compilation with ontologies.
\newblock In: B.~Pfahringer, J.~Renz (eds.) Proceedings of the 28th
  Australasian Joint Conference on Advances in Artificial Intelligence (AI'15),
  \emph{LNCS}, vol. 9457, pp. 457--463. Springer (2015).
\newblock Canberra, ACT, Australia, November 30 -- December 4, 2015

\bibitem{Pan10}
Pan, W.L., Liu, D.x.: Mapping object role modeling into common logic
  interchange format.
\newblock In: Proceedings of the 3rd International Conference on Advanced
  Computer Theory and Engineering (ICACTE'10), vol.~2, pp. 104--109. IEEE
  Computer Society (2010)

\bibitem{Parent06}
Parent, C., Spaccapietra, S., Zim\'{a}nyi, E.: Conceptual modeling for
  traditional and spatio-temporal applications---the MADS approach.
\newblock Berlin Heidelberg: Springer Verlag (2006)

\bibitem{Partridge13}
Partridge, C., Gonzalez-Perez, C., Henderson-Sellers, B.: Are conceptual models
  concept models?
\newblock In: W.~Ng, V.C. Storey, J.~Trujillo (eds.) 32nd International
  Conference on Conceptual Modeling (ER'13), \emph{LNCS}, vol. 8217, pp.
  96--105. Springer (2013).
\newblock 11-13 November, 2013, Hong Kong

\bibitem{Queralt12}
Queralt, A., Artale, A., Calvanese, D., Teniente, E.: {OCL-Lite}: Finite
  reasoning on {UML/OCL} conceptual schemas.
\newblock Data \& Knowledge Engineering \textbf{73}, 1--22 (2012)

\bibitem{Queralt08}
Queralt, A., Teniente, E.: Decidable reasoning in {UML} schemas with
  constraints.
\newblock In: Z.~Bellahsene, M.~L{\'e}onard (eds.) Proceedings of the 20th
  International Conference on Advanced Information Systems Engineering
  (CAiSE'08), \emph{LNCS}, vol. 5074, pp. 281--295. Springer (2008).
\newblock Montpellier, France, June 16-20, 2008

\bibitem{Safwat16}
Safwat, H., Davis, B.: {CNLs} for the semantic web: a state of the art.
\newblock Language Resources \& Evaluation \textbf{51}(1), 191--220 (2017)

\bibitem{Shoval97}
Shoval, P., Shiran, S.: Entity-relationship and object-oriented data
  modeling---an experimental comparison of design quality.
\newblock Data and Knowledge Engineering \textbf{21}, 297--315 (1997)

\bibitem{Smaragdakis09}
Smaragdakis, Y., Csallner, C., Subramanian, R.: Scalable satisfiability
  checking and test data generation from modeling diagrams.
\newblock Automation in Software Engineering \textbf{16}, 73--99 (2009)

\bibitem{Solanki16}
Solanki, M., Brewster, C.: {OntoPedigree:} modelling pedigrees for traceability
  in supply chains.
\newblock Semantic Web Journal \textbf{7}(5), 483--491 (2016)

\bibitem{Song09}
Song, I.Y., Chen, P.P.: Entity relationship model.
\newblock In: L.~Liu, M.T. \"Ozsu (eds.) Encyclopedia of Database Systems,
  vol.~1, pp. 1003--1009. Springer (2009)

\bibitem{Soylu17}
Soylu, A., Kharlamov, E., Zheleznyakov, D., Ruiz, E.J., Giese, M., Skjaeveland,
  M.G., Hovland, D., Schlatte, R., Brandt, S., Lie, H., Horrocks, I.:
  Optique{VQS}: a visual query system over ontologies for industry.
\newblock Semantic Web Journal \textbf{9}(5), 627--660 (2018)

\bibitem{Thalheim09}
Thalheim, B.: Extended entity relationship model.
\newblock In: L.~Liu, M.T. \"Ozsu (eds.) Encyclopedia of Database Systems,
  vol.~1, pp. 1083--1091. Springer (2009)

\bibitem{Tobies01}
Tobies, S.: Complexity results and practical algorithms for logics in knowledge
  representation.
\newblock Ph.D. thesis, RWTH Aachen (2001)

\bibitem{toman2009applications}
Toman, D., Weddell, G.E.: Applications and extensions of {PTIME} {D}escription
  {L}ogics with functional constraints.
\newblock In: Proceedings of the 21st International Joint Conference on
  Artificial Intelligence {IJCAI'09}, pp. 948--954. AAAI Press (2009)

\bibitem{Toman11}
Toman, D., Weddell, G.E.: Fundamentals of Physical Design and Query
  Compilation.
\newblock Synthesis Lectures on Data Management. Morgan {\&} Claypool (2011)

\bibitem{TW14}
Toman, D., Weddell, G.E.: On adding inverse features to the description logic
  \emph{CFD}\({}^{\mbox{{\(\forall\)}}}\)\({}_{\mbox{nc}}\).
\newblock In: {PRICAI} 2014: Trends in Artificial Intelligence - 13th Pacific
  Rim International Conference on Artificial Intelligence, Gold Coast, QLD,
  Australia, December 1-5, 2014., pp. 587--599 (2014)

\bibitem{Venable95}
Venable, J., Grundy, J.: Integrating and supporting {E}ntity {R}elationship and
  {O}bject {R}ole {M}odels.
\newblock In: M.P. Papazoglou (ed.) Proceedings of the 14th International
  Conference on Object-Oriented and Entity-Relationship Modelling (ER'95),
  \emph{LNCS}, vol. 1021, pp. 318--328. Springer (1995).
\newblock Gold Coast, Australia, December 12-15, 1995

\bibitem{Wagih13}
Wagih, H.M., Zanfaly, D.S.E., Kouta, M.M.: Mapping {Object Role Modeling 2}
  schemes into $\mathcal{SROIQ}(d)$ description logic.
\newblock International Journal of Computer Theory and Engineering
  \textbf{5}(2), 232--237 (2013)

\bibitem{West10}
West, M., Partridge, C., Lycett, M.: Enterprise data modelling: Developing an
  ontology-based framework for the shell downstream business.
\newblock In: R.~Cuel, R.~Ferrario (eds.) Proceedings of Formal Ontologies Meet
  industry (FOMI'10), pp. 71--84 (2010).
\newblock 14-15 December 2010, Trento, Italy

\end{thebibliography}
\end{document}